\RequirePackage[T1]{fontenc}
\documentclass[10pt,journal,compsoc]{IEEEtran}
\usepackage[utf8]{inputenc}
\usepackage{cite}
\usepackage{amsmath,amssymb}
\usepackage{graphicx}
\usepackage{algorithm}
\usepackage{algpseudocode}
\usepackage{booktabs}
\usepackage{multirow}
\usepackage{xcolor}

\usepackage{subcaption}
\usepackage{hyperref}
\usepackage{xurl}
\usepackage{placeins}

\hypersetup{hidelinks}
\graphicspath{{figures/}{./}}

\algrenewcommand\algorithmicrequire{\textbf{Input:}}
\algrenewcommand\algorithmicensure{\textbf{Output:}}

\begin{document}

\title{Less Is More in the Long Tail: Stage-Adaptive Sample Selection for Annotation-Efficient Dense Prediction}

\author{Xiaofei~Du,
        Lei~Zhang,
        Shuyu~Yan,
        Manning~Wang,
        and~Zhijian~Song%
\thanks{X. Du, S. Yan, M. Wang, and Z. Song are with the Digital Medical Research Center, School of Basic Medical Sciences, Fudan University, Shanghai 200032, China.}%
\thanks{X. Du, S. Yan, M. Wang, and Z. Song are also with the Shanghai Key Laboratory of Medical Image Computing and Computer Assisted Intervention, Shanghai 200032, China.}%
\thanks{X. Du, S. Yan, M. Wang, and Z. Song are also with the Frontier Innovation Center, Department of Systems Biology for Medicine, Qidong-Fudan Innovative Institute of Medical Sciences, School of Basic Medical Sciences, Zhongshan Hospital Clinical Center for Biotherapy, Fudan University, Shanghai 200032, China.}%
\thanks{L. Zhang is with the Department of Computer Science, University of Exeter, Exeter EX4 4QF, United Kingdom.}%
\thanks{Corresponding authors: Manning Wang and Zhijian Song (e-mail: mnwang@fudan.edu.cn; zjsong@fudan.edu.cn).}}

\markboth{IEEE Transactions on Pattern Analysis and Machine Intelligence}%
{Du \MakeLowercase{\textit{et al.}}: SASS}

\IEEEtitleabstractindextext{%
\begin{abstract}
Deep learning performance generally improves with increasing training data, yet this scaling is fundamentally constrained by annotation cost in large-scale dense prediction tasks with long-tailed category distributions, where pixel- or voxel-level annotation is prohibitively expensive. We propose SASS (Stage-Adaptive Sample Selection), a stage-adaptive data-selection framework for pool-based active learning in long-tailed dense prediction. SASS combines three components: label-free self-supervised gradient scoring, prior-guided category rebalancing with validation-driven feedback, and stage-adaptive acquisition aligned with model training dynamics. This design avoids candidate ground-truth masks during gradient scoring while making acquisition responsive to long-tail imbalance and evolving representations. We evaluate SASS on a multimodal 3D medical segmentation testbed comprising over 100{,}000 samples spanning 108 anatomical structures. SASS recovers 98.3\% of full-dataset performance with a 40\% training-pool annotation budget, outperforming BADGE by 5.1 percentage points. Moreover, SASS exhibits a statistically supported ``less-is-more'' pattern, surpassing full-dataset training at the Hard-group level and, at the structure level, for the pancreas and gallbladder. More broadly, SASS shows that annotation-efficient learning depends not only on which samples are selected, but also on how the annotation budget is distributed across categories and when model-derived scores begin to guide selection.
\end{abstract}

\begin{IEEEkeywords}
Active learning, annotation-efficient learning, data selection, dense prediction, long-tailed learning, medical image segmentation.
\end{IEEEkeywords}
} 

\maketitle

\section{Introduction}
\label{sec:introduction}

Deep learning has become a fundamentally data-driven learning paradigm, particularly in supervised settings where model performance scales strongly with the volume and quality of labeled data. While self-supervised pretraining reduces dependence on labels for representation learning, high performance on downstream tasks---especially dense prediction at the pixel or voxel level---still relies critically on extensive supervised annotation~\cite{kirillov2023sam,ma2024medsam,wang2023sam_med3d}. As training corpora grow to hundreds of thousands of samples spanning fine-grained categories, annotation increasingly dominates the resource budget, shifting the central question from \emph{how much data to collect} to \emph{which data to annotate}.

Data selection broadly concerns choosing a subset of available data to improve task performance under resource constraints~\cite{coleman2020selection,mirzasoleiman2020coresets}. When annotation is the constrained resource, this objective is commonly formulated as pool-based active learning: a model repeatedly selects examples for labeling from an available set of unlabeled candidates~\cite{settles2009active,ren2021survey}. Acquisition functions---rules that score or prioritize candidates for annotation---guide these decisions under a fixed labeling budget~\cite{ash2020badge}. The objective is to approach the performance of full-dataset training while annotating only a small fraction of the candidate pool.

Existing approaches address sample selection from complementary perspectives---including gradient- and influence-based valuation, uncertainty estimation, and diversity-driven acquisition. Gradient-, influence-, and data-valuation methods estimate sample value through parameter gradients, validation effects, or marginal contribution to model performance~\cite{koh2017understanding,pruthi2020estimating,yeh2018representer,ghorbani2019data,killamsetty2021gradmatch}, highlighting their reliance on supervised candidate or validation signals. Uncertainty- and information-based acquisition functions score samples by predictive entropy, confidence, or expected information gain~\cite{settles2009active,gal2017deep,kirsch2019batchbald}, while diversity- and coverage-based methods select representative samples in feature space~\cite{sener2018active,coleman2020selection}; both lines often operate without explicit category-level rebalancing in long-tailed settings. Moreover, representative acquisition functions are typically defined as fixed scoring criteria~\cite{settles2009active,sener2018active,ash2020badge}, despite evidence that example difficulty, gradients, and learned representations evolve substantially during training~\cite{toneva2019forgetting,jacot2018neural,koh2017understanding}. Despite decades of research on active learning and data selection, three coupled challenges remain insufficiently addressed. These challenges become increasingly severe as supervised learning scales to large, long-tailed datasets---in which a few categories have many examples while many others have relatively few---with expensive annotations.

Collectively, these assumptions give rise to three fundamental limitations. First, \emph{supervision dependency}: a broad class of gradient- and influence-based valuation methods requires a supervised loss gradient for each candidate, whether that gradient is scored directly or used to estimate the candidate's effect on loss over a labeled validation set. Computing such a gradient requires the candidate's ground-truth mask, which is unavailable before acquisition. Existing approaches partially bypass this requirement using pseudo-labels or expected losses~\cite{ash2020badge,wang2022boosting}, but the resulting gradient estimates remain dependent on the current model's predictions. Second, \emph{category-imbalance bias}: under long-tailed data distributions, category-agnostic acquisition functions, which do not explicitly balance selections across categories, can disproportionately allocate annotation budget to abundant categories because they dominate gradient and uncertainty statistics~\cite{settles2009active,sener2018active,ash2020badge}, systematically under-selecting rare or difficult categories where additional supervision may be most beneficial. For example, the pancreas illustrates this problem: it occupies only a small fraction of an abdominal volume, has low-contrast boundaries, and may nevertheless be the structure of primary clinical interest. Third, \emph{training-dynamics mismatch}: most acquisition methods apply a fixed selection rule throughout training, implicitly assuming that a single notion of sample informativeness remains appropriate as optimization progresses. This is problematic for deep networks, whose representations and gradients evolve substantially; consequently, a criterion that is informative at one stage may not remain equally informative at another~\cite{toneva2019forgetting,achille2018critical}. The combination of long-tailed data and costly annotation is not specific to medical imaging: it arises broadly in large-scale supervised learning~\cite{cui2019class,liu2019large,cao2019learning,kang2020decoupling,settles2009active,sener2018active,coleman2020selection,killamsetty2021glister}.

Among such large-scale, annotation-constrained learning problems, volumetric medical image segmentation provides a particularly challenging setting in which these three limitations arise simultaneously. Dense annotation of a single 3D volume can require hours of specialist effort ~\cite{wang2024comprehensive,budd2021survey}, making exhaustive labeling impractical at scale. Meanwhile, anatomical structures vary markedly in size, visual complexity, segmentation difficulty, and sample availability~\cite{wasserthal2023totalseg}, creating strongly heterogeneous and long-tailed supervision. These challenges are further compounded for rare and difficult structures by low tissue contrast, irregular boundaries, and limited spatial context~\cite{isensee2021nnu}, making segmentation errors both more likely and clinically more consequential. Volumetric medical image segmentation therefore provides a representative and demanding setting for studying principled data selection under severe annotation constraints.

To address these challenges, we propose \textbf{SASS (Stage-Adaptive Sample Selection)}, a stage-adaptive data-selection framework for pool-based active learning in long-tailed dense prediction. Rather than treating acquisition only as sample ranking, SASS jointly controls how the annotation budget is distributed across categories and when model-derived gradients begin to rank individual candidates. It addresses the three limitations jointly through label-free self-supervised gradient valuation (scoring candidates without their ground-truth masks), long-tail-aware category rebalancing, and validation-gated activation of gradient scoring, implemented through the following three principles. First, \emph{label-free self-supervised gradient scoring} measures candidate informativeness using a self-supervised teacher--student gradient computed between the current model, used as a frozen teacher, and a previous checkpoint, used as the student. The resulting gradient norm quantifies how strongly each unlabeled candidate induces representation change across training states. Second, \emph{prior-guided rebalancing} incorporates category-level priors---instantiated in our medical setting as scale-aware and coverage-balancing priors---and further refines them through validation-driven feedback that converts per-category performance gaps into acquisition priorities. Third, \emph{stage-adaptive acquisition} adapts the selection criterion to the evolving reliability of model-derived gradient signals. During the initial stage, category-level priors guide selection and candidate gradients are not computed. Once the predefined validation criterion is met, gradient scoring is activated with a progressively increasing weight, while the prior terms retain nonzero weights to preserve category coverage.

\subsection{Contributions}

Our main contributions are as follows: 
\begin{enumerate}

    \item  We introduce a \textbf{self-supervised acquisition scorer} that ranks unlabeled candidates via teacher-student DINO gradient norms, eliminating the need for ground-truth masks during candidate scoring and thereby addressing the supervision dependency of conventional gradient- and influence-based acquisition.

    \item We design a prior-guided, validation-driven rebalancing mechanism that combines scale-aware and coverage-balance priors with category-level validation feedback to prioritize underperforming long-tail categories. A group-budget constraint further prevents redundant category groups from dominating the acquisition budget. This reallocation yields a ``less-is-more'' pattern: with only 40\% of the annotation budget, SASS matches or exceeds full-dataset training on several challenging structures and achieves a statistically supported gain for the Hard group (Section~\ref{sec:difficulty_results}).

    \item We propose a stage-adaptive acquisition strategy that addresses the training-dynamics mismatch of fixed selection rules by adapting acquisition to the evolving reliability of selection signals during optimization. SASS emphasizes prior-guided exploration in early training, when model-derived evidence is less reliable, and transitions to gradient-driven refinement as learned representations mature and self-supervised influence estimates become more informative.

    \item We provide \textbf{large-scale empirical evidence} for annotation-efficient sample selection in 3D dense prediction. We evaluate SASS on a multimodal 3D corpus of over 100{,}000 samples spanning 108 anatomical structures. The underlying segmentation model is trained entirely from random initialization, without pretrained weights or external checkpoints. With a 40\% training-pool annotation budget, SASS recovers 98.3\% of full-dataset performance, outperforms BADGE by 5.1\,pp, and demonstrates a statistically supported ``less-is-more'' effect on hard categories.

\end{enumerate}
\section{Related Work}

\subsection{Active Learning and Data Selection}
\label{sec:al_selection}

Data selection has become a central challenge in large-scale deep learning because training samples contribute unequally to model learning, motivating methods that prioritize informative subsets~\cite{paul2021deep,mindermann2022prioritized,mirzasoleiman2020coresets,killamsetty2021glister,killamsetty2021gradmatch}. Active learning focuses on the annotation-acquisition setting by selecting informative samples from an unlabeled pool under a fixed labeling budget~\cite{settles2009active}. Classical acquisition functions can be broadly grouped into uncertainty-, diversity-, and gradient-based strategies, each exposing limitations in long-tailed dense-prediction regimes.

\emph{Uncertainty-based methods} query samples with high predictive entropy, small prediction margins, or large epistemic uncertainty~\cite{settles2009active,gal2017deep}. Scoring candidates independently can yield redundant batches; BatchBALD addresses this problem by considering their joint information about model parameters~\cite{kirsch2019batchbald}. Learning Loss trains an auxiliary module on labeled examples to predict the losses of unlabeled candidates~\cite{yoo2019}. These scores prioritize predictive difficulty or information gain but do not explicitly enforce balanced category coverage.

\emph{Diversity-based methods} seek geometric coverage of the data distribution. Coreset selection formulates acquisition as a $k$-center problem~\cite{sener2018active}, and VAAL uses an adversarially trained latent space to identify underrepresented unlabeled samples~\cite{sinha2019vaal}. These methods reduce redundancy by improving coverage of the candidate distribution. More recently, Sel4FT performs one-shot annotation selection for pretrained models by jointly preserving distributional fidelity to the unlabeled pool and sample diversity~\cite{lu2025sel4ft}. However, geometric coverage or distribution preservation alone does not guarantee balanced annotation coverage when the candidate pool is long-tailed.

\emph{Hybrid and gradient-based methods} combine informativeness and diversity or estimate sample value from model gradients.
BADGE forms gradient embeddings using predicted labels and applies $k$-means++ batch selection~\cite{ash2020badge}.
For existing training sets, gradient-preserving coreset methods approximate full-data optimization with selected subsets~\cite{mirzasoleiman2020coresets,killamsetty2021gradmatch}.
LESS uses gradient similarity to select instruction-tuning examples relevant to a target task~\cite{xia2024less}. Diff-In further improves influence estimation by accumulating influence differences across training steps~\cite{tan2026diffin}. These methods provide strong model-dependent signals. For valuation scores defined by supervised candidate losses, application to an unlabeled pool requires predicted targets or another explicitly specified surrogate objective. Trajectory-aware valuation improves how such scores are estimated over training, but does not by itself determine whether a model-derived signal should enter an acquisition rule that also accounts for long-tail category coverage.

RALF learns a time-varying exploration--exploitation trade-off from classifier feedback, with its acquisition criteria remaining available throughout the process~\cite{ebert2012ralf}.

Beyond the design of acquisition criteria, two additional lines of work are particularly relevant to our setting: long-tailed rebalancing and data-efficient dense prediction.

\emph{Long-tailed learning and rebalancing.} Long-tailed learning typically intervenes through training objectives, classifiers, or pseudo-labels. Representative directions include re-weighting~\cite{cui2019class}, margin-based losses~\cite{cao2019learning}, and decoupled representation and classifier learning~\cite{kang2020decoupling}. Recent studies offer a unified view of loss-oriented methods~\cite{wang2026unified}, address unknown test label distributions~\cite{yang2026dirmixe}, and handle mismatched labeled and unlabeled distributions~\cite{gan2026decon}. These approaches improve learning under a given data distribution, but do not determine which samples should be annotated to reshape the labeled-pool composition under a limited budget. Imbalance-aware acquisition has also been explored through acquisition-time balancing for imbalanced classification and class-balanced region selection for 2D semantic segmentation~\cite{aggarwal2020active,cai2021revisiting}. For example, Cai et al. combine class-balanced superpixel selection with click-based annotation-cost measurement~\cite{cai2021revisiting}.

\emph{Active learning and data selection for dense prediction.} Large-scale dense prediction further amplifies the limitations of existing acquisition functions. In volumetric medical segmentation, annotations are expensive, category distributions are long-tailed, and the same acquisition rule may behave differently as representations evolve during training~\cite{budd2021survey,wang2024comprehensive}. Outside medical imaging, recent work has addressed data efficiency in dense prediction through point-efficient active learning for salient object detection~\cite{wu2025pixel}, replay-sample selection for continual semantic segmentation~\cite{zhu2025replaymaster}, and multi-modal semi-supervised learning for 3D scene understanding~\cite{kong2025lasermix}. Related active-learning methods include warm-start selection with proxy labels~\cite{nath2022warm}, feature-mixing strategies for diverse batch construction~\cite{parvaneh2022active}, and graph-based sequential selection~\cite{caramalau2021sequential}. However, many are developed for 2D or task-specific settings and do not directly address label-free self-supervised gradient scoring, category-level long-tail rebalancing, and stage-dependent acquisition in large-scale 3D dense prediction.

\subsection{Self-Supervised Acquisition Signals and Training Dynamics}
\label{sec:ssl_gradient_curriculum}

Self-supervised learning and teacher--student models have been combined with active learning to reduce labeling effort, for example through self-training on unlabeled data or knowledge-distillation-based uncertainty estimation~\cite{bengar2021ssl_al,peng2021lane_kd_al}. In the classification experiments of Bengar et al., active querying offered little additional benefit after self-supervised pretraining at low labeling budgets, whereas the combination was beneficial at higher budgets~\cite{bengar2021ssl_al}. Peng et al. estimate image uncertainty from knowledge learned by a student model and combine it with a diversity criterion for lane detection~\cite{peng2021lane_kd_al}.

Beyond how acquisition signals are constructed, their utility may also change as training progresses: a sample that is informative at one stage may become less useful later, and vice versa. Gradient statistics can support one-shot pruning of already annotated data when label-dependent scores are averaged over multiple initializations~\cite{paul2021deep}; however, these conditions do not transfer directly to an unlabeled candidate pool, and a score computed once does not account for subsequent changes in the model or selected pool. Data selection is therefore closely related to curriculum design: effective learning depends not only on which samples are selected, but also on when and how they are presented during training. Curriculum learning organizes training from easier to harder examples~\cite{bengio2009curriculum}, and self-paced learning formalizes this idea as joint optimization over model parameters and sample weights~\cite{kumar2010self}. A central challenge is defining difficulty, because static measures such as loss or entropy may not capture how sample informativeness changes during training. Forgetting events---transitions from correct to incorrect predictions---provide a dynamic measure of example difficulty and have been used to distinguish consistently remembered, frequently forgotten, and persistently difficult samples~\cite{toneva2019forgetting}.

For dense prediction, this stage dependence also affects how curriculum and forgetting signals should be defined. A single volume may contain both easy and hard categories, while rare structures provide sparse prediction histories. Pool expansion further changes batch composition, coupling curriculum design to sample selection. Curriculum strategies have used medically informed difficulty measures for fracture classification~\cite{jimenez2019medical}, while class-rebalancing self-training preferentially adds minority-class pseudo-labels during training~\cite{wei2021crest}. However, their use as a stabilization mechanism coupled with active sample selection in large-scale 3D multi-organ training remains less studied.

\subsection{Large-Scale Medical Segmentation and Annotation Cost}
\label{sec:foundation_models}

Medical image segmentation has progressed from task-specific systems such as nnU-Net~\cite{isensee2021nnu} and transformer-based volumetric architectures~\cite{hatamizadeh2022unetr,tang2022self} to larger multi-organ and promptable models. AbdomenCT-1K~\cite{ma2022abdomenct1k}, TotalSegmentator~\cite{wasserthal2023totalseg}, AMOS~\cite{ji2022amos}, and AbdomenAtlas~\cite{qu2024abdomenatlas} expanded multi-organ coverage, while SAM~\cite{kirillov2023sam}, MedSAM~\cite{ma2024medsam}, SegVol~\cite{du2023segvol}, SAM-Med3D~\cite{wang2023sam_med3d}, SAM-Med3D-MoE~\cite{wang2024sammed3dmoe}, and CAT~\cite{gao2024cat} advanced promptable or foundation-style segmentation. These models increase the demand for heterogeneous, densely annotated data across many anatomical structures and modalities. The resulting annotation bottleneck makes data selection a central scaling problem rather than a peripheral preprocessing step.

These advances broaden segmentation capability, but do not by themselves determine which additional volumes should be annotated under a limited labeling budget.
\section{Method}
\label{sec:method}

\begin{figure*}[!t]
    \centering
    \begin{subfigure}[b]{0.97\textwidth}
        \centering
        \includegraphics[width=\textwidth]{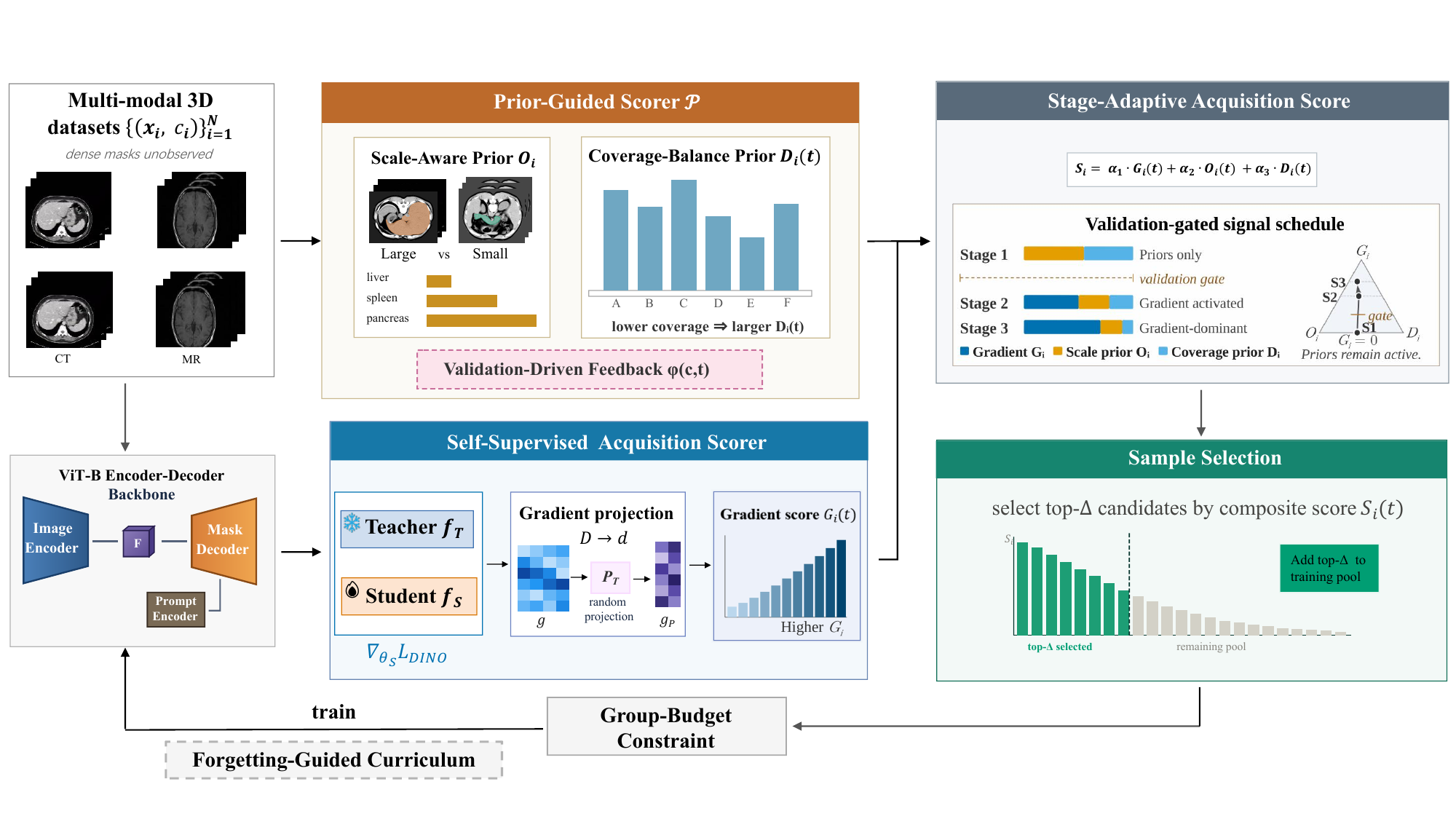}
        \caption{Overview of the SASS framework}
        \label{fig:framework_a}
    \end{subfigure}
    \vspace{0.5em}
    \begin{subfigure}[b]{0.97\textwidth}
        \centering
        \includegraphics[width=\textwidth]{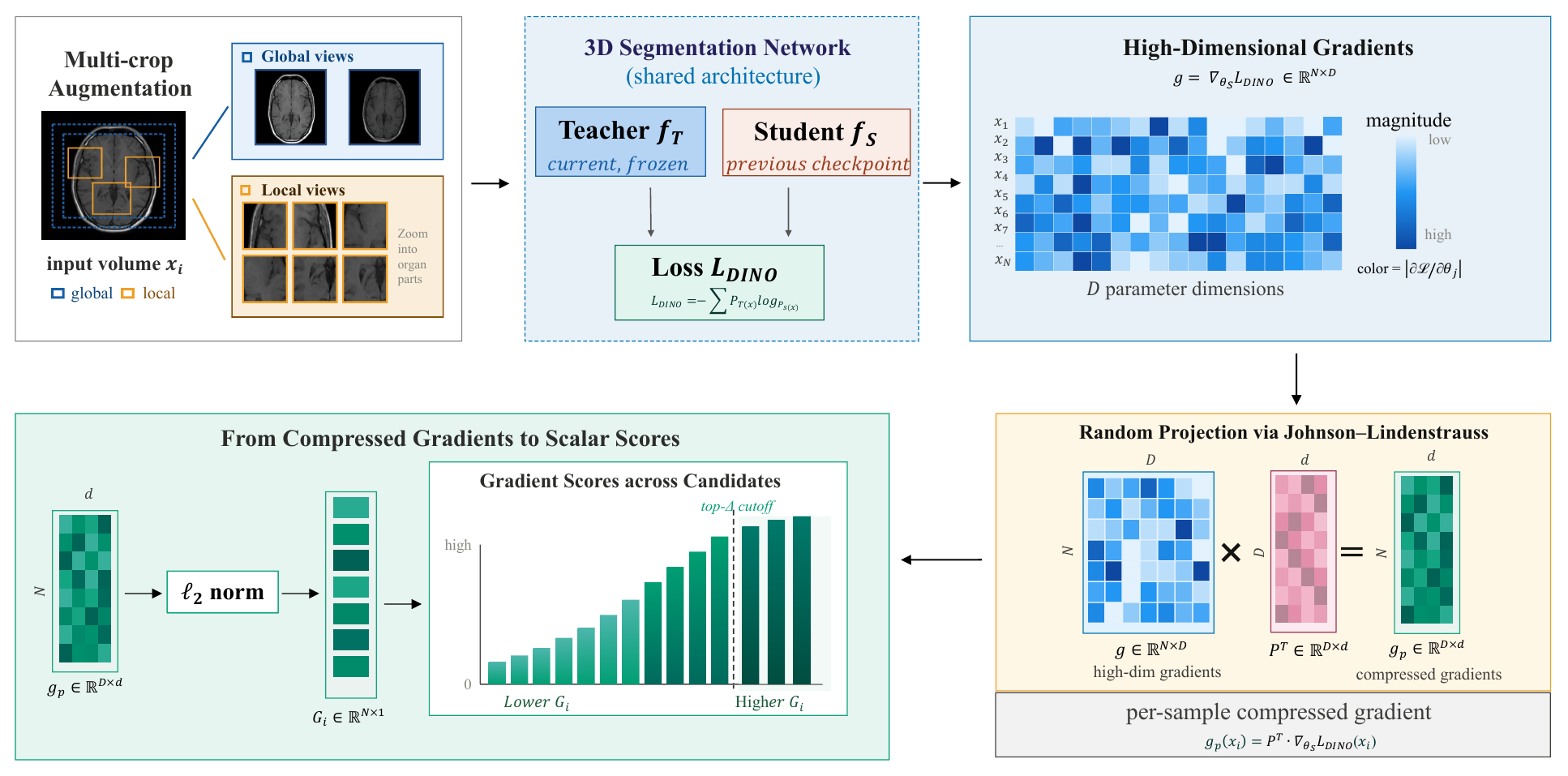}
        \caption{Gradient computation pipeline with dimension reduction}
        \label{fig:framework_b}
    \end{subfigure}
    \caption{Overview of SASS. The training pool grows progressively from $|\mathcal{S}^{(0)}|\approx0.10N$ to the target budget through periodic expansion. At each selection round, the \emph{Self-Supervised Acquisition Scorer} computes label-free gradient scores $G_i(t)$ from a teacher-student DINO objective, while the \emph{Prior-Guided Scorer} produces a \emph{Scale-Aware Prior} $O_i$ and a \emph{Coverage-Balance Prior} $D_i(t)$, both enhanced by \emph{Validation-Driven Feedback} $\phi(c,t)$. The \emph{Stage-Adaptive Acquisition Score} normalizes and combines these signals into $S_i(t)$, with weights shifting from prior-dominant exploration to gradient-dominant refinement. In the ternary simplex, each stage is placed at its weight vector, the three vertices denoting full weight on $G_i$, $O_i$ and $D_i$; Stage~1 lies on the $O_i$--$D_i$ edge, where the gradient weight is zero, whereas Stages~2 and~3 lie inside the triangle, so both priors keep non-zero weight. Selected samples are refined by a \emph{Group-Budget Constraint} and reordered by a \emph{Forgetting-Guided Curriculum} before training.}
    \label{fig:framework}
\end{figure*}

\subsection{Overview}
\label{sec:method_overview}

Given a candidate pool of image-category pairs $\{(x_i,c_i)\}_{i=1}^{N}$ whose dense masks are initially unobserved, our goal is to select a subset $\mathcal{S}$ of size $|\mathcal{S}|=\rho N$ for annotation so that training on $\mathcal{S}$ approaches full-dataset performance with substantially lower labeling cost. SASS addresses the changing relative utility of acquisition signals over optimization by assigning greater weight to category-level priors early in training and progressively increasing the contribution of self-supervised gradients at later stages.

As illustrated in Fig.~\ref{fig:framework}, each selection round evaluates candidates using three complementary signals. The \emph{Self-Supervised Acquisition Scorer} computes a label-free gradient score $G_i(t)$ that measures the strength of the self-supervised update induced by candidate $x_i$. The \emph{Prior-Guided Scorer} $\mathcal{P}$ produces two category-level priors: a \emph{Scale-Aware Prior} $O_i$, instantiated by organ-size statistics in our medical testbed, and a \emph{Coverage-Balance Prior} $D_i(t)$, which penalizes categories already over-represented in the selected pool. These prior signals are further enhanced by \emph{Validation-Driven Feedback} $\phi(c,t)$, which increases acquisition priority for categories whose validation performance lags behind the global median. The \emph{Stage-Adaptive Acquisition Score} normalizes and fuses the gradient, scale, and coverage signals into a composite score $S_i(t)$ using time-varying weights that shift from prior-dominant exploration in early training to gradient-dominant refinement in later stages.

SASS further includes two mechanisms that stabilize selection and training. A \emph{Group-Budget Constraint} limits redundant parent groups such as ribs and vertebrae in our medical instantiation, and a \emph{Forgetting-Guided Curriculum} reorders selected samples from easy to hard based on prediction-stability history. Finally, an adaptive pool-growth schedule expands the annotated set through fixed-size query batches. A performance gate may defer scheduled expansions during validation decline, producing variable consolidation intervals, while validation-driven reselection refreshes acquisition priorities as the model matures.

\subsection{Self-Supervised Gradient-Based Sample Scoring}
\label{sec:gradient_scoring}

The \emph{Self-Supervised Acquisition Scorer} takes an unlabeled candidate pool $\mathcal{U}$ and two model snapshots---the current model $\theta^{(t)}$ and a previous checkpoint $\theta^{(t_{\mathrm{prev}})}$---and outputs a scalar score $G_i(t)$ for each candidate $x_i\in\mathcal{U}$. Conventional supervised influence estimators often rely on supervised candidate gradients, which require candidate masks before annotation. We instead compute gradients of a self-supervised DINO objective~\cite{caron2021dino}, allowing candidate scoring without ground-truth masks.

The gradient computation pipeline, illustrated in Fig.~\ref{fig:framework}(b), consists of three sequential stages: (1)~a teacher-student architecture that computes the self-supervised DINO loss for each candidate (Section~\ref{sec:gradient_compute}), (2)~random projection that compresses the resulting high-dimensional gradients to a tractable dimensionality (Section~\ref{sec:projection}), and (3)~gradient norm computation that converts the compressed gradient vector into a scalar gradient score (Section~\ref{sec:grad_score}). We describe each stage below.

\subsubsection{Teacher-Student Knowledge Evolution}
\label{sec:gradient_compute}

At each selection round (epoch $t$), we construct two model snapshots to obtain a candidate-dependent self-supervised update signal across training states. The teacher $f_T=\mathrm{freeze}(\theta^{(t)})$ is a frozen copy of the current model, whereas the student $f_S=\theta^{(t_{\mathrm{prev}})}$ is restored from the previous selection state. This teacher-student asymmetry allows SASS to measure how strongly each candidate induces a self-supervised update across two stages of training.

For the first gradient-based selection round, the student is loaded from a checkpoint $\omega$ epochs earlier, establishing the initial temporal gap. In subsequent rounds, the student is restored from the model snapshot saved at the preceding expansion round. Let $t_k$ denote the epoch of the $k$-th realized expansion; the teacher--student temporal gap is then $\tau_k=t_k-t_{k-1}$ and may vary when the performance gate defers an expansion. Gradient-based scoring is activated only after the validation-based condition in Eq.~\ref{eq:activation} is satisfied.

This design estimates self-supervised knowledge evolution over the interval $[t_{\mathrm{prev}},t]$. A large gradient indicates that candidate $x_i$ induces a strong teacher-student update under the DINO objective, making it a label-free proxy for acquisition value without requiring ground-truth masks.

\paragraph{Multi-scale crop strategy}
Following the multi-crop augmentation introduced in DINO~\cite{caron2021dino}, we generate multi-scale views from each candidate volume $x_i$: $n_g$ global crops with scale range $[\sigma_g^{\min},\sigma_g^{\max}]$ and $n_l$ local crops with scale range $[\sigma_l^{\min},\sigma_l^{\max}]$. In dense prediction tasks with large intra-dataset scale variation---in our testbed, anatomical structures range from large organs to small glands---global crops preserve holistic spatial context, whereas local crops emphasize fine structural variation and boundary detail. This multi-scale design makes the teacher-student discrepancy less dependent on a single crop scale.

\paragraph{Representation learning with DINO loss}
Teacher and student process all crops through the network, including the image encoder, prompt encoder, and mask decoder. The decoder outputs are flattened and projected into a $d_p$-dimensional space using a fixed Gaussian random projection matrix $\mathbf{P}_{\mathrm{DINO}}\in\mathbb{R}^{D_{\mathrm{dec}}\times d_p}$. The DINO loss for sample $x_i$ is
\begin{equation}
    \mathcal{L}_{\mathrm{DINO}}(x_i)
    =
    -\sum_{v\in\mathcal{V}_g}
    \sum_{\substack{v'\in\mathcal{V}_g\cup\mathcal{V}_l\\v'\neq v}}
    \mathbf{p}_T(v)\cdot\log\mathbf{p}_S(v'),
    \label{eq:dino_loss}
\end{equation}
where $\mathcal{V}_g$ and $\mathcal{V}_l$ denote global and local crop sets. Following DINO~\cite{caron2021dino}, $\mathbf{p}_T(v)$ and $\mathbf{p}_S(v)$ are softmax distributions over the projected decoder outputs $\mathbf{P}_{\mathrm{DINO}}^\top\mathbf{z}_T(v)$ and $\mathbf{P}_{\mathrm{DINO}}^\top\mathbf{z}_S(v)$, respectively, with temperatures $\tau_T<\tau_S$ and a centered EMA term $\mathbf{c}$ with momentum $\beta_c$ applied to the teacher output to reduce representation collapse.

\paragraph{Knowledge evolution gradient}
For each candidate sample $x_i$, the gradient with respect to the student parameters is
\begin{equation}
    \mathbf{g}_i=\nabla_{\theta_S}\mathcal{L}_{\mathrm{DINO}}(x_i).
    \label{eq:gradient}
\end{equation}
Since $\nabla_{\theta_S}\mathcal{L}_{\mathrm{DINO}}$ is driven by the teacher-student discrepancy, $\|\mathbf{g}_i\|_2$ quantifies the magnitude of the parameter-space update induced by candidate $x_i$ under the self-supervised objective. Importantly, this gradient is computed without ground-truth labels, enabling model-dependent candidate scoring without ground-truth masks. The resulting high-dimensional gradient vector $\mathbf{g}_i\in\mathbb{R}^{D}$ is then passed to the compression stage described next.

\subsubsection{Random Projection for Gradient Compression}
\label{sec:projection}

The per-sample gradient $\mathbf{g}_i\in\mathbb{R}^{D}$ computed in Section~\ref{sec:gradient_compute} resides in the full parameter space of the ViT-B encoder ($D\approx93$M). At this dimensionality, explicitly storing and comparing gradients across a large candidate pool incurs substantial memory and computational overhead. To make gradient-based acquisition scalable, we compress each gradient into a lower-dimensional representation using a Rademacher random projection, which approximately preserves the pairwise geometry of the original gradient space:

\begin{equation}
    \tilde{\mathbf{g}}_i=\mathbf{P}^{\top}\mathbf{g}_i,\qquad
    \mathbf{P}\in\mathbb{R}^{D\times d},
    \label{eq:projection}
\end{equation}
where $P_{jk}\in\{-1/\sqrt{d},+1/\sqrt{d}\}$ and $d\ll D$.

By the Johnson-Lindenstrauss lemma~\cite{johnson1984extensions}, random projection approximately preserves pairwise distances among candidate gradients, so the relative geometry used for sample ranking is largely maintained after compression. The Rademacher construction provides an efficient binary projection with lower computational cost than dense Gaussian projections~\cite{achlioptas2003database}. After projection, this reduces gradient storage from $O(nD)$ to $O(nd)$ and permits $O(nd)$ norm-based scoring over $n$ candidates.

\subsubsection{Gradient Norm as Acquisition Score}
\label{sec:grad_score}

Given the compressed gradient $\tilde{\mathbf{g}}_i\in\mathbb{R}^{d}$, we define the self-supervised gradient score as:
\begin{equation}
    G_i(t)=\|\tilde{\mathbf{g}}_i\|_2.
    \label{eq:grad_score}
\end{equation}
This score quantifies the magnitude of the parameter update induced by candidate $i$ under the current teacher-student representation, providing a model-dependent measure of its potential contribution to learning.

Conventional gradient-based data selection typically estimates sample influence through directional similarity, such as the dot product between candidate and validation gradients~\cite{xia2024less,koh2017understanding}. Such formulations are not directly applicable to our setting for two reasons. First, computing supervised candidate gradients requires candidate masks that are unavailable before acquisition, even though a labeled validation set is available. Second, under the shared DINO teacher-student objective, candidate gradients are driven toward a common representation-alignment objective, producing a substantial shared directional component across samples. Directional similarity can therefore be dominated by this common component rather than reflecting differences in candidate informativeness. We instead use gradient magnitude to measure how strongly each unlabeled candidate perturbs the current model under the self-supervised objective. A larger norm indicates a greater model update is required to reconcile the candidate with the teacher representation, whereas a small norm indicates that the current representation already explains the candidate with relatively little adjustment. Gradient magnitude thus provides a label-free, model-dependent acquisition signal that remains discriminative without requiring an external validation gradient or relying on directional variation between candidates.

\textbf{Remark~1} (Interpretation of gradient norm under DINO loss).
Let $z_i$ denote the student representation used by the DINO head. By the chain rule,
\begin{equation}
    \mathbf{g}_i
    = \nabla_{\theta_S}\mathcal{L}_{\mathrm{DINO}}(x_i)
    = J_{\theta_S}(x_i)^\top\delta_i,
    \label{eq:chain_rule}
\end{equation}
where $\delta_i=\partial\mathcal{L}_{\mathrm{DINO}}/\partial z_i$ measures the teacher-student discrepancy at the student output, and $J_{\theta_S}(x_i)=\partial z_i/\partial\theta_S$ measures the sensitivity of the student representation to its parameters. Thus, $\|\mathbf{g}_i\|_2$ scores candidates whose self-supervised discrepancy induces a strong parameter-space update. This retains the parameter-update interpretation while requiring no ground-truth masks for candidate samples. Because the random projection approximately preserves gradient norms, $G_i(t)=\|\tilde{\mathbf{g}}_i\|_2$ retains this parameter-update interpretation after compression.

\subsection{Prior-Guided Sample Scoring}
\label{sec:prior_scoring}

Long-tailed dense prediction datasets contain categories whose annotation value is not captured by frequency alone. Category-agnostic acquisition functions, including uncertainty sampling~\cite{settles2009active} and BADGE~\cite{ash2020badge}, may therefore under-select rare or difficult categories when the candidate pool is highly imbalanced. In our medical testbed, this issue appears as anatomical heterogeneity: organ volumes span orders of magnitude, sample availability is uneven across structures, and segmentation difficulty varies with tissue contrast, boundary complexity, and spatial context. We therefore introduce a \emph{Prior-Guided Scorer} that encodes two category-level priors---a \emph{Scale-Aware Prior} and a \emph{Coverage-Balance Prior}---and further adapts them using validation-driven feedback.

\subsubsection{Scale-Aware Prior}
\label{sec:scale_prior}

Because candidate masks are unavailable during selection, SASS cannot use candidate-specific volumes. Instead, it employs a pre-specified category-level scale prior available before acquisition. In our medical instantiation, the known category identifier $c_i$ is used only to index a metadata lookup table, and the expected organ volume is defined as $\hat{V}_i=\mu_{c_i}$, where $\mu_c$ is the reference mean volume associated with category $c$. No candidate-specific volume or ground-truth mask is accessed when computing this prior.

The \emph{Scale-Aware Prior} inversely weights the expected category scale:
\begin{equation}
O_i = \frac{1}{(\hat{V}_i)^\gamma / \lambda_v + 1}
\label{eq:organ_size}
\end{equation}
where $\gamma$ controls the sensitivity to scale differences and $\lambda_v$ is a scaling constant. The score decreases monotonically with expected structure volume, thereby increasing the relative priority of small structures. This is particularly relevant in volumetric segmentation, where small structures often provide fewer foreground voxels and less spatial context for learning and are more susceptible to class imbalance and boundary errors~\cite{isensee2021nnu}. The prior therefore encodes a stable, annotation-free preference toward categories for which additional supervision may be particularly valuable, rather than relying on unavailable candidate-level geometry.

\subsubsection{Coverage-Balance Prior}
\label{sec:coverage_prior}

Scale alone does not account for how the annotation budget has already been distributed. As acquisition proceeds, repeatedly selecting candidates from the same categories can lead to redundant allocation while leaving other categories insufficiently represented. SASS therefore complements the static scale prior with a dynamic \emph{Coverage-Balance Prior} that adapts to the composition of the selected pool. Let $\mathcal{H}^{(t)}$ denote the samples selected up to epoch $t$. The \emph{Coverage-Balance Prior} is
\begin{equation}
D_i(t) = \frac{1}{1 + \log(1 + n_{c_i}^{(t)})},
\label{eq:diversity}
\end{equation}
where $n_{c_i}^{(t)} = |\{j \in \mathcal{H}^{(t)} : \mathrm{cat}(j) = c_i\}|$. The logarithmic penalty encourages exploration of under-represented categories while avoiding excessive penalization of categories that have already been sampled.

\subsubsection{Validation-Driven Feedback}
\label{sec:validation_feedback}

Static priors capture dataset-level structure but cannot reflect the model's evolving category-wise performance during training. If selection relies only on fixed priors, it may continue sampling categories that are already well learned while under-sampling categories that remain difficult. We therefore introduce validation-driven feedback to convert category-level validation gaps into acquisition priorities for subsequent selection rounds.

\paragraph{Data source and timing}
Since the candidate pool is unlabeled, category-wise performance cannot be measured from candidates directly. Feedback is therefore derived exclusively from the validation set $\mathcal{V}$. During training, the model is periodically evaluated on held-out validation samples, producing category-wise Dice scores $\{\mathrm{Dice}_c^{(t)}\}_{c=1}^{C}$. These scores are computed before the next selection round and cached by the training loop. When a selection round is triggered, the cached scores are used to compute $\phi(c,t)$ (Eq.~\ref{eq:gap}), which then enhances the scale and coverage priors into $O_i^*(t)$ and $D_i^*(t)$ for ranking unlabeled candidates. This temporal ordering---evaluate the current model on validation data, then use feedback to score the unlabeled pool---ensures that candidate labels are never used for scoring.

Let $\widetilde{\mathrm{Dice}}^{(t)}=\mathrm{median}\bigl(\{\mathrm{Dice}_c^{(t)}\}_{c=1}^{C}\bigr)$ denote the median category Dice at epoch $t$. The feedback signal for category $c$ is
\begin{equation}
    \phi(c,t)=\frac{2}{1+\exp\!\bigl(-\kappa\cdot\max(0,\widetilde{\mathrm{Dice}}^{(t)}-\mathrm{Dice}_c^{(t)})\bigr)}-1,
    \label{eq:gap}
\end{equation}
where $\kappa$ is a sharpness parameter. The signal is zero for categories at or above the median and increases monotonically for underperforming categories. The sigmoid scaling keeps the feedback smooth and bounded, preventing extreme score inflation.

\paragraph{Score enhancement}
The scale and coverage priors are enhanced as
\begin{align}
    O_i^*(t) &= \mathrm{clip}\!\left(O_i + \lambda^{(s)} \cdot \phi(c_i, t),\; 0,\; 1\right) \label{eq:enhanced_organ} \\
    D_i^*(t) &= \mathrm{clip}\!\left(D_i(t) \cdot \bigl(1 + \mu^{(s)} \cdot \phi(c_i, t)\bigr),\; 0,\; 1\right) \label{eq:enhanced_diversity}
\end{align}

where $\lambda^{(s)}$ and $\mu^{(s)}$ are enhancement strengths for stage $s\in\{2,3\}$. Validation-driven feedback is inactive during Stage~1, which uses the unenhanced priors $O_i$ and $D_i(t)$; the initial pool is therefore constructed from dataset-level structure alone, before any model-derived signal enters selection. Categories with low validation Dice receive elevated scores for new candidates from the unlabeled pool, increasing their exposure in subsequent acquisition rounds. Concrete values of $\kappa$, $\lambda^{(s)}$, and $\mu^{(s)}$ are specified in Section~\ref{sec:baselines} and ablated in Section~\ref{sec:ablation}.

\subsubsection{Group-Budget Constraint}
\label{sec:capping}

The Scale-Aware Prior $O_i$ (Eq.~\ref{eq:organ_size}) and Coverage-Balance Prior $D_i(t)$ (Eq.~\ref{eq:diversity}) regulate acquisition at the individual-category level, based on expected scale and accumulated selection frequency, respectively. However, category-level balancing does not necessarily imply balanced allocation at a coarser semantic level. When a dataset contains many related categories within the same parent group, each category can appear individually under-represented while their aggregate allocation becomes disproportionately large. This creates a hierarchical imbalance that cannot be identified from per-category statistics alone.

This issue is particularly pronounced for repetitive anatomical structures such as ribs and vertebrae. Although individual ribs or vertebrae are treated as distinct segmentation categories, they share substantial anatomical and visual characteristics. Allocating additional annotations to each category independently can therefore yield diminishing marginal benefit while consuming a large fraction of the total annotation budget. Validation-driven feedback can modulate category priorities according to current performance, but it does not explicitly control how much of the budget is collectively assigned to a related group. We therefore introduce a \emph{Group-Budget Constraint} that imposes an explicit upper bound on the acquisition share of predefined repetitive groups, complementing the soft category-level scoring terms.

\paragraph{Why soft scoring alone is insufficient}
Soft category-level scoring cannot fully resolve redundancy when many related categories belong to the same parent group. From the perspective of Eq.~\ref{eq:diversity}, selecting ``rib\_01'' does not penalize ``rib\_02'' because they are formally distinct categories, even though adjacent ribs share similar morphology, contrast profiles, and boundary characteristics. The logarithmic penalty also saturates slowly, allowing each repetitive category to accumulate selections while the parent group dominates the pool. Validation-driven feedback $\phi(c,t)$ (Eq.~\ref{eq:gap}) can deprioritize well-performing categories, but it may also boost a few underperforming repetitive categories in early training. In preliminary experiments without hard capping, the ribs+vertebrae group consumed over 80\% of the annotation budget despite the combined effect of coverage scoring and validation feedback.

\paragraph{Hard capping as a group-level constraint}
We introduce a \emph{Group-Budget Constraint} that limits the fraction of each query batch assigned to a predefined parent group:
\begin{equation}
    \begin{gathered}
    \bigl|\{i \in \Delta_k : \mathrm{group}(i)=g\}\bigr|
    \leq \lceil \eta |\Delta_k| \rceil,\\
    g \in \{\mathrm{rib},\mathrm{vertebrae}\},
    \end{gathered}
    \label{eq:super_cap}
\end{equation}
where $\Delta_k$ is the query batch at expansion round $k$. We also limit any single category:
\begin{equation}
    \bigl|\{i \in \Delta_k : \mathrm{cat}(i)=c\}\bigr|
    \leq \lceil \eta_s |\Delta_k| \rceil.
    \label{eq:single_cap}
\end{equation}

The budget released by capping is reassigned to the highest-scoring remaining candidates. Thus, the soft priors in Eqs.~\ref{eq:diversity}--\ref{eq:enhanced_diversity} provide relative rebalancing across categories, whereas Eqs.~\ref{eq:super_cap}--\ref{eq:single_cap} impose absolute group-level budget guarantees. The cap ratios $\eta$ and $\eta_s$ are specified in Section~\ref{sec:baselines}. Group-cap sensitivity is analyzed in Section~\ref{sec:ablation}.

\subsection{Stage-Adaptive Acquisition}
\label{sec:stage_adaptive}

The effectiveness of data selection depends on aligning acquisition criteria with the non-stationary learning dynamics of deep networks~\cite{bengio2009curriculum,coleman2020selection,mindermann2022prioritized}. Early gradient statistics can support effective one-shot data pruning~\cite{paul2021deep}; however, their rankings need not remain equally suitable throughout successive acquisition rounds as task-specific representations evolve. Category-level priors provide a gradient-independent basis for broad coverage early in training, while gradient-based selection can play an increasing role as task-specific representations develop and training dynamics evolve~\cite{toneva2019forgetting,hacohen2019power}. Motivated by this shift in relative utility, SASS dynamically calibrates the contribution of gradient, scale, and coverage signals over training.

\subsubsection{Training Stages and Transition Conditions}
\label{sec:stages}

We partition the $T_{\max}$-epoch training process into three stages corresponding to different levels of model maturity. Each stage uses a different mixture of acquisition signals, and the transition to gradient-inclusive selection is conditioned on validation performance. Fig.~\ref{fig:framework}(a) visualizes this schedule as a trajectory in the simplex of admissible weight vectors $\{(\alpha_1,\alpha_2,\alpha_3):\alpha_k\geq0,\ \sum_k\alpha_k=1\}$, where a point's distance from the edge opposite a vertex is proportional to the weight placed on the corresponding signal. Stage~1 is confined to the $O_i$--$D_i$ edge, so the validation gate marks a departure from that edge rather than a gradual re-weighting of three signals that are active throughout.

\paragraph{Stage~1: Prior-guided exploration (epochs $0$ to $T_1$)}
During early training, gradient-based rankings may remain sensitive to initialization and may not yet reflect stable task-specific semantics. From an NTK perspective, gradient relations near initialization are governed by the initialization-induced kernel while task-specific features are still developing~\cite{jacot2018neural}. We therefore set the gradient weight to zero and rely on prior-guided scoring to establish broad category coverage:

\begin{equation}
    \mathbf{w}_1=(0,\beta_O^{(1)},\beta_D^{(1)}),\qquad
    \beta_O^{(1)}+\beta_D^{(1)}=1.
    \label{eq:stage1_weights}
\end{equation}
This avoids premature reliance on gradient-based rankings, while an elevated scale-prior weight $\beta_O^{(1)}$ emphasizes small categories and the coverage-prior weight $\beta_D^{(1)}$ discourages early over-representation.

\paragraph{Stage~2: Performance-conditioned transition (epochs $T_1$ to $T_2$)}

As training progresses, task-specific representations become more stable and gradient geometry becomes increasingly structured, providing a stronger basis for gradient-based candidate ranking before full model convergence~\cite{xia2024less,coleman2020selection}. SASS therefore activates gradient-inclusive selection only after the model reaches a validation-performance threshold:

\begin{equation}
    \mathrm{Activate\ Stage~2}\iff
    (t\geq T_1)\wedge
    \bigl(\mathrm{Dice}_{\mathrm{val}}^{(t)}\geq\tau_{\mathrm{act}}\bigr).
    \label{eq:activation}
\end{equation}

Activation is latched: once the condition in Eq.~\ref{eq:activation} is satisfied at some epoch, gradient scoring remains enabled for the remainder of training even if validation Dice subsequently fluctuates below $\tau_{\mathrm{act}}$.

Upon activation, the weights interpolate linearly over $[T_1,T_2]$:
\begin{equation}
    \alpha_k^{(2)}(t)
    =\alpha_k^{\mathrm{start}}
    +\frac{t-T_1}{T_2-T_1}
    \left(\alpha_k^{\mathrm{end}}-\alpha_k^{\mathrm{start}}\right),
    \label{eq:stage2_interp}
\end{equation}
where $k\in\{1,2,3\}$ and $\sum_k\alpha_k=1$. During this transition, the gradient weight $\alpha_1$ increases while the prior weights $\alpha_2$ and $\alpha_3$ decrease, smoothly shifting acquisition from prior-guided exploration to gradient-guided refinement.

\paragraph{Stage~3: Gradient-dominant refinement (epochs $T_2$ to $T_{\max}$)}
Weights are fixed at gradient-dominant values:
\begin{equation}
    \mathbf{w}_3=(\alpha_1^{(3)},\alpha_2^{(3)},\alpha_3^{(3)}).
    \label{eq:stage3_weights}
\end{equation}
The dominant gradient weight prioritizes samples with large knowledge-evolution scores (Section~\ref{sec:grad_score}), while the reduced but nonzero prior weights preserve baseline category coverage and prevent selection from collapsing onto a narrow high-gradient subset.

All stage hyperparameters ($T_1$, $T_2$, $\tau_{\mathrm{act}}$, all $\alpha$ and $\beta$ values) are specified in Section~\ref{sec:training_config} and ablated in Section~\ref{sec:ablation}.

\subsubsection{Composite Acquisition Score}

Because neither model-driven gradients nor prior-based signals alone provide a reliable acquisition criterion across all training stages, SASS combines them into a stage-adaptive composite score:

\begin{equation}
    S_i(t)=\alpha_1(t)\hat{G}_i(t)+\alpha_2(t)\hat{O}_i^*(t)+\alpha_3(t)\hat{D}_i^*(t),
    \label{eq:composite_score}
\end{equation}
where $\hat{G}_i(t)$, $\hat{O}_i^*(t)$, and $\hat{D}_i^*(t)$ are min-max normalized to $[0,1]$:
\begin{equation}
    \hat{X}_i=\frac{X_i-\min_j X_j}{\max_j X_j-\min_j X_j+\epsilon},
\end{equation}
where $\epsilon$ prevents division by zero.

This schedule implements a time-varying trade-off: early selection emphasizes prior-guided exploration, whereas late selection emphasizes gradient-guided refinement. Activation is by design discontinuous: the gradient weight steps from zero to $\alpha_1^{\mathrm{start}}$ once the validation criterion is met, reflecting that gradient evidence is either admitted or withheld rather than gradually trusted. The linear interpolation thereafter avoids further abrupt changes within Stage~2.

\subsubsection{Adaptive Scheduling Algorithm}

The per-round selection pipeline is summarized in Supplementary Algorithm~S1. The stage-adaptive weight computation follows Eqs.~\ref{eq:stage1_weights}--\ref{eq:stage3_weights}; the complete pseudocode is provided in the supplementary material.

The stage-adaptive design also reduces computational overhead by activating gradient-based acquisition only when the corresponding model-derived signal becomes sufficiently informative. Because $\alpha_1=0$ in Stage~1, SASS bypasses gradient computation during the first $T_1$ epochs and further postpones gradient scoring when the validation-performance threshold in Eq.~\ref{eq:activation} has not yet been reached.

\subsection{Forgetting-Guided Curriculum}
\label{sec:curriculum}

The \emph{Forgetting-Guided Curriculum} is designed to stabilize optimization after each pool expansion rather than to serve as an additional acquisition criterion. Newly annotated query batches change the composition of the training set and can introduce transient instability when difficult samples are presented before the model has adapted to the expanded pool. Motivated by curriculum and self-paced learning~\cite{bengio2009curriculum,kumar2010self}, SASS estimates sample difficulty from prediction histories and orders selected samples from easier to harder cases within each training epoch.

\subsubsection{Forgetting Events}

A forgetting event for sample $i$ at epoch $t$ is a transition from a correct to an incorrect prediction:
\begin{equation}
    \mathrm{FE}_i^{(t)}
    =\mathbb{1}\!\left[\mathrm{correct}_i^{(t-1)}\wedge
    \neg\mathrm{correct}_i^{(t)}\right].
    \label{eq:forgetting_event}
\end{equation}
For dense segmentation, a prediction is considered correct when its Dice score exceeds the adaptive threshold $\tau_{\mathrm{corr}}^{(t)}=\mathrm{clip}(\overline{\mathrm{Dice}}_{\mathrm{train}}^{(t)}-\xi,\tau_{\mathrm{corr}}^{\min},\tau_{\mathrm{corr}}^{\max})$, where $\overline{\mathrm{Dice}}_{\mathrm{train}}^{(t)}$ is the running mean training Dice over the labeled training samples and $\xi$ is a margin below that mean. The cumulative forgetting count available at epoch $t$ is $F_i=\sum_{\tau=1}^{t}\mathrm{FE}_i^{(\tau)}$.

\subsubsection{Difficulty Estimation and Curriculum Ordering}

Let $H_i$ denote the recorded prediction history of sample $i$, and let $r_i=|\{t:\mathrm{correct}_i^{(t)}\}|/|H_i|$ be its correctness ratio. A forgetting count is informative only when a sample has been predicted correctly at least occasionally: a persistently incorrect sample may have $F_i=0$ despite being difficult or atypical. We therefore partition the selected pool into a rankable set $\mathcal{S}_{\mathrm{rank}}=\{i\in\mathcal{S}:r_i\geq r_{\min}\}$ and a deferred set $\mathcal{S}_{\mathrm{defer}}=\mathcal{S}\setminus\mathcal{S}_{\mathrm{rank}}$.

For each sample in $\mathcal{S}_{\mathrm{rank}}$, the curriculum difficulty is
\begin{equation}
    d_i=\min\!\left(1,F_i\zeta_i+d_{\mathrm{base}}\right),
    \qquad i\in\mathcal{S}_{\mathrm{rank}},
    \label{eq:difficulty_score}
\end{equation}
where $\zeta_i=\zeta_{\mathrm{sv}}$ when $r_i>r_{\mathrm{sv}}$ and $\zeta_i=\zeta_{\mathrm{def}}$ otherwise, with $\zeta_{\mathrm{def}}<\zeta_{\mathrm{sv}}$. Repeatedly forgotten but otherwise learnable samples consequently receive larger difficulty scores, consistent with prior observations on example-forgetting dynamics~\cite{toneva2019forgetting}.

At each epoch, rankable samples are ordered from easy to hard, while persistently incorrect samples are deferred rather than misclassified as easy:
\begin{equation}
    \mathcal{S}_{\mathrm{ordered}}
    =
    \operatorname{argsort}_{i\in\mathcal{S}_{\mathrm{rank}}}d_i
    \;\Vert\;
    \operatorname{Shuffle}(\mathcal{S}_{\mathrm{defer}}),
    \label{eq:curriculum_order}
\end{equation}
where $\Vert$ denotes sequence concatenation. Deferred samples remain available
for training but are presented after the forgetting-ranked samples. This ordering changes sample presentation within the labeled pool; it does not acquire additional labels.

\subsection{Periodic Incremental Pool Growth}
\label{sec:pool_growth}

Rather than allocating the full annotation budget $\rho N$ at initialization, SASS expands the selected pool over multiple acquisition rounds so that selection can adapt to the model's evolving representations.

\subsubsection{Consolidation-Aligned Expansion Schedule}
Training-stage perturbations can affect subsequent optimization dynamics~\cite{achille2018critical}. Each newly annotated query batch changes the composition of the training set. If successive expansions occur before the model has adapted to the preceding batch, the optimizer state and learned representations may remain temporarily misaligned with the expanded pool, leading to transient training instability. Moreover, gradient scores computed immediately after expansion may reflect short-term adaptation to the recent pool change rather than persistent candidate value. SASS therefore separates acquisition rounds by a consolidation interval.

The initial pool is defined by the initial-pool ratio $\rho_0$:
\begin{equation}
    |\mathcal{S}^{(0)}|\approx \rho_0 N,\qquad 0<\rho_0\leq\rho,
    \label{eq:initial_pool}
\end{equation}
where $N$ is the size of the candidate pool and $\rho$ is the target annotation ratio. The initial pool is selected using Stage~1 prior-guided scoring. SASS performs $K_\Delta$ additive expansions. At the $k$-th realized expansion epoch $t_k$, a query batch of $\Delta$ newly annotated samples is added:
\begin{equation}
    |\mathcal{S}^{(k)}|
    =|\mathcal{S}^{(k-1)}|+\Delta,\qquad
    k=1,\ldots,K_\Delta,
    \label{eq:pool_growth}
\end{equation}
until the target budget $\rho N$ is reached. Thus, $\mathcal{S}^{(k-1)}\subset\mathcal{S}^{(k)}$, and all previously selected samples remain available for subsequent training. The realized consolidation interval is
\begin{equation}
    \tau_k=t_k-t_{k-1},
    \label{eq:realized_interval}
\end{equation}
which need not be constant because the performance gate may defer an expansion.

Two safeguards regulate pool growth. First, a \emph{Performance Gate} defers a scheduled expansion when recent validation Dice satisfies the predefined decline criterion. Second, \emph{Validation-Triggered Rescoring} recomputes candidate scores when validation progress stagnates. Rescoring does not itself expand the annotated pool; instead, it updates the candidate ranking used at the next expansion. The training and evaluation setup is described in Section~\ref{sec:training_config}.

\section{Experiments}
\label{sec:experiments}

\subsection{Experimental Setup}
\label{sec:experimental_setup}

\subsubsection{Datasets and Splits}
\label{sec:datasets}

The candidate pool aggregates publicly available 3D segmentation datasets, principally TotalSegmentator~\cite{wasserthal2023totalseg}, AMOS~\cite{ji2022amos}, BraTS21~\cite{baid2021brats}, KiTS~\cite{heller2021kits}, MMWHS~\cite{zhuang2016mmwhs}, CrossMoDA~\cite{dorent2023crossmoda}, FLARE22~\cite{ma2023flare22}, WORD~\cite{luo2022word}, AbdomenCT-1K~\cite{ma2022abdomenct1k}, and VerSe~\cite{sekuboyina2021verse}. Dataset aggregation and label harmonization follow Wang et al.~\cite{wang2023sam_med3d}; only publicly accessible data are used. The complete 108-structure mapping, dataset licenses, and access procedures are provided in Supplementary Tables~S1--S2.

Each acquisition unit is an image--category pair $(x_i,c_i)$ with one category-specific binary mask that remains unobserved until the unit is selected. Volumes with multi-class annotations are decomposed into per-category masks following the point-prompt protocol of Wang et al.~\cite{wang2023sam_med3d}. The resulting pool contains 100{,}357 units across 108 structures, spanning CT and multiple MRI sequences, and is stratified by category into 70{,}351 training candidates, 15{,}003 validation samples, and 15{,}003 held-out test samples. The validation set is used to compute category-level feedback $\phi(c,t)$, trigger reselection, and select SASS hyperparameters; the held-out test set is never accessed during acquisition, model selection, or hyperparameter tuning and is used only for final performance reporting. A stricter volume-disjoint robustness check is reported in Table~\ref{tab:volume_disjoint}; the split construction and re-evaluation protocol are described in Supplementary Section~S3.

\subsubsection{Training and Evaluation}
\label{sec:training_config}

We use the point-prompt 3D segmentation architecture of Wang et al.~\cite{wang2023sam_med3d}, comprising a 93M-parameter ViT-B image encoder, a 3D prompt encoder, and a 3D mask decoder. All parameters are trained from random initialization; no pretrained weights or external checkpoints are loaded. SASS operates at the data-selection level and requires no architectural modification (Section~\ref{sec:generalization}).

Volumes are resampled to $128^3$ (trilinear for images, nearest-neighbor for masks) with foreground Z-score intensity normalization. Augmentation details are in Supplementary Section~S4. Optimization uses AdamW ($\beta_1{=}0.9$, $\beta_2{=}0.999$, weight decay $0.05$) with base learning rate $8{\times}10^{-4}$ for the encoder and $0.1{\times}$ for the prompt encoder and decoder, 5-epoch warmup, and cosine annealing over $T_{\max}{=}300$ epochs. The loss is equally weighted Dice and cross-entropy. The effective batch size is 480 on two NVIDIA RTX 6000-class GPUs. Training uses a single pseudo-click; validation and testing use 10-click iterative refinement~\cite{wang2023sam_med3d}. Within each run, overall Dice is the unweighted macro-average of the 108 structure-level Dice means; difficulty-group Dice averages the structures in the corresponding fixed group. Reported means and standard deviations are computed across three independent runs.

For SASS-versus-Full comparisons, we use 10{,}000 paired hierarchical bootstrap replicates over runs and source volumes, and summarize each difference by its bootstrap 95\% confidence interval. $\dagger$: interval excludes zero; $\ddagger$: positive mean difference whose interval includes zero. Per-structure intervals are not adjusted for multiplicity; the Hard-group comparison in Section~\ref{sec:difficulty_results} is a single pre-specified test.

\subsubsection{Selection Protocol and Baselines}
\label{sec:baselines}

The training pool is treated as unlabeled; masks are revealed only after selection. The annotation budget counts selected training image--category pairs and excludes the labeled validation set used for acquisition feedback and hyperparameter selection. The principal budget is $\rho{=}0.40$ (${\sim}$28{,}000 samples), starting from $\rho_0{=}0.10$ and expanding in batches of $\Delta{=}3{,}000$. SASS uses $(T_1,T_2){=}(40,150)$, $\tau_{\mathrm{act}}{=}0.42$, $d{=}2{,}048$, $\kappa{=}5$, $\eta{=}0.15$, $\eta_s{=}0.08$. Stage weights: Stage~1 $(\beta_O^{(1)},\beta_D^{(1)}){=}(0.55,0.45)$; Stage~2 interpolates $\alpha_1$ from 0.30 to 0.70; Stage~3 fixes $(\alpha_1,\alpha_2,\alpha_3){=}(0.70,0.20,0.10)$. Feedback strengths: $(\lambda^{(2)},\mu^{(2)}){=}(0.35,0.50)$, $(\lambda^{(3)},\mu^{(3)}){=}(0.45,0.75)$. Additional implementation details are provided in Supplementary Section~S4.

Baselines: Random, Entropy Sampling~\cite{settles2009active}, Coreset~\cite{sener2018active}, BADGE~\cite{ash2020badge}, LESS~\cite{xia2024less}, and LESS+Organ (LESS with a static Scale-Aware Prior). Full Dataset and Full+CB (inverse-frequency sampler) use 100\%. All subset methods share the same 40\% budget, acquisition protocol, backbone, optimizer, and evaluation protocol. All SASS-specific hyperparameters are selected exclusively on the validation set. Weighted-sampler, feature-extraction, and distributed-training details are provided in Supplementary Section~S5.

\subsection{Main Results}
\label{sec:results}

\subsubsection{Difficulty-Stratified Performance}
\label{sec:difficulty_results}

To analyze performance as a function of segmentation difficulty, anatomical structures are grouped according to the validation Dice achieved by the Full-Dataset baseline. This provides a SASS-independent reference for segmentation difficulty while avoiding bias from the proposed selection method itself. Specifically, structures are grouped by the Full Dataset validation Dice: Easy (${\geq}0.70$, $n{=}59$), Medium ($[0.55,0.70)$, $n{=}31$), Hard ($<0.55$, $n{=}18$). Assignments are fixed before test-set evaluation (Supplementary Section~S7).

\begin{table*}[!t]
\centering
\caption{Difficulty-stratified test Dice at epoch~300 under 10-click evaluation (mean $\pm$ std, three runs). Subset methods use 40\%; Full and Full+CB use 100\%. Bold: best 40\%-budget result.}
\label{tab:main_difficulty}
\scriptsize
\setlength{\tabcolsep}{2.8pt}
\begin{tabular}{@{}lcccccccccc@{}}
\toprule
Group & Random & Entropy & Coreset & BADGE & LESS & LESS+O & SASS & Full & Full+CB & $\Delta$(S--R) \\
\midrule
Easy ($n{=}59$)   & 0.651$\pm$0.003 & 0.665$\pm$0.003 & 0.661$\pm$0.003 & 0.674$\pm$0.003 & 0.674$\pm$0.003 & 0.691$\pm$0.003 & \textbf{0.715$\pm$0.003} & 0.733$\pm$0.002 & 0.732$\pm$0.003 & +0.064 \\
Medium ($n{=}31$) & 0.596$\pm$0.004 & 0.610$\pm$0.004 & 0.605$\pm$0.004 & 0.618$\pm$0.004 & 0.618$\pm$0.004 & 0.641$\pm$0.004 & \textbf{0.675$\pm$0.004} & 0.690$\pm$0.003 & 0.693$\pm$0.004 & +0.079 \\
Hard ($n{=}18$)   & 0.525$\pm$0.006 & 0.538$\pm$0.006 & 0.534$\pm$0.006 & 0.549$\pm$0.005 & 0.553$\pm$0.005 & 0.578$\pm$0.005 & \textbf{0.622$\pm$0.005} & 0.610$\pm$0.005 & 0.618$\pm$0.005 & +0.097 \\
\midrule
Overall (108)     & 0.614$\pm$0.003 & 0.628$\pm$0.003 & 0.624$\pm$0.003 & 0.637$\pm$0.003 & 0.638$\pm$0.003 & 0.658$\pm$0.003 & \textbf{0.688$\pm$0.003} & 0.700$\pm$0.002 & 0.702$\pm$0.003 & +0.074 \\
\bottomrule
\end{tabular}
\end{table*}

Table~\ref{tab:main_difficulty} shows that SASS recovers 98.3\% of Full Dataset Dice using 40\% of annotations, improving over Random by 7.4\,pp, over BADGE by 5.1\,pp, and over LESS+Organ by 3.0\,pp. The gain increases with difficulty: on Hard structures, SASS exceeds Full by 1.2\,pp (95\% CI: $[0.003,0.021]$, $p{=}0.011$), supporting the group-level \emph{less-is-more} effect analyzed in Section~\ref{sec:selection_evolution}.

\begin{figure*}[!t]
\centering
\includegraphics[width=0.95\textwidth]{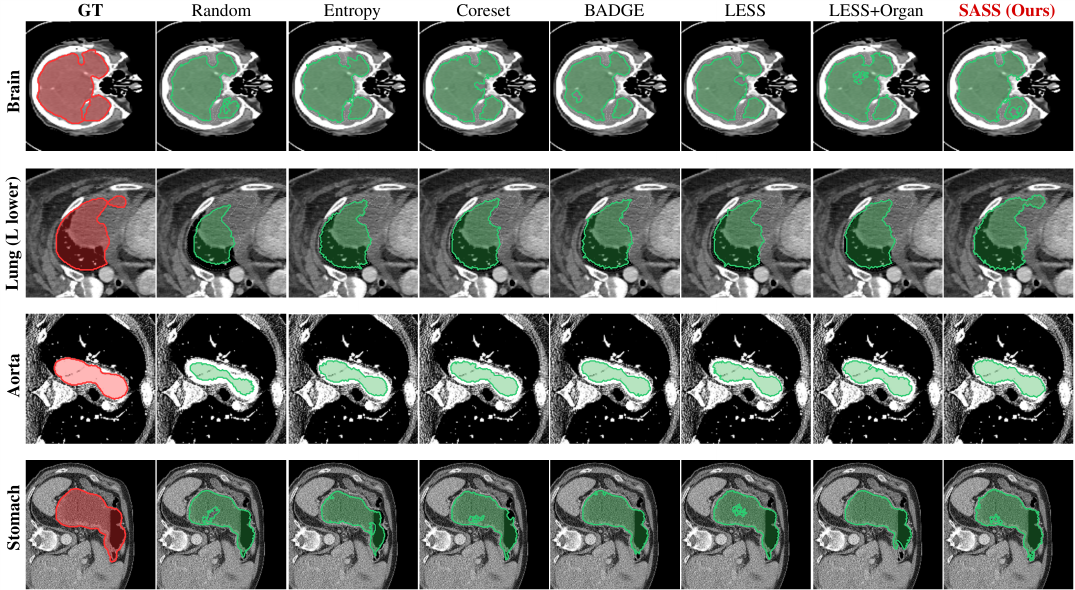}
\caption{Qualitative segmentation comparison on four anatomical structures. Red: ground truth; green: prediction; dashed white: ground-truth boundary. LESS+O denotes LESS+Organ.}
\label{fig:segmentation}
\end{figure*}

Figure~\ref{fig:segmentation} shows that SASS more closely follows ground-truth boundaries, particularly for the Aorta and Stomach where baselines exhibit visible under-segmentation.

\subsubsection{Comparison with Active Learning Methods}
\label{sec:al_comparison}

All subset-selection methods outperform Random in Table~\ref{tab:main_difficulty}, confirming substantial redundancy in the candidate pool. Among the classical active-learning acquisition functions (Entropy, Coreset, BADGE), BADGE achieves the highest overall Dice (0.637); LESS and LESS+Organ are influence-based data-selection methods rather than acquisition functions and are therefore reported as a separate reference group. The diagnostic LESS+Organ variant adds a static Scale-Aware Prior to the LESS gradient score but uses no validation feedback, no group constraints, and a fixed acquisition rule; it reaches 0.658. SASS provides a further 3.0\,pp gain, attributable to validation feedback, stage-adaptive weighting, and group-budget control (Section~\ref{sec:ablation}).

To separate informative selection from category rebalancing, we compare with Full+CB. Full+CB raises Hard-structure Dice from 0.610 to 0.618 using 100\% of annotations; SASS reaches 0.622 with only 40\%. Its overall Dice remains 1.4\,pp below Full+CB (0.688 vs.\ 0.702), indicating that SASS's advantage over full-supervision references is concentrated on the long tail rather than uniformly distributed across categories.

\subsubsection{High-Priority Structure Performance}
\label{sec:clinical_organs}

\begin{table}[!t]
\centering
\caption{Dice for 18 high-priority reporting targets under the 10-click protocol at epoch~300. Rows include bilateral and lung-lobe aggregates and report mean $\pm$ std across three runs; the final row is the unweighted mean of the 18 displayed row means. Targets were selected by diagnostic relevance~\cite{wasserthal2023totalseg}. $\dagger$/$\ddagger$: see Section~\ref{sec:training_config}.}
\label{tab:high_priority}
\scriptsize
\setlength{\tabcolsep}{3.5pt}
\begin{tabular}{@{}lcccc@{}}
\toprule
Structure & Rand & Full & SASS & $\Delta$(S--R) \\
\midrule
Liver              & .701$\pm$.006 & .793$\pm$.005 & .781$\pm$.005          & +.080 \\
Kidney (L/R)       & .688$\pm$.007 & .776$\pm$.005 & .764$\pm$.006          & +.076 \\
Spleen             & .695$\pm$.007 & .782$\pm$.005 & .771$\pm$.006          & +.076 \\
Pancreas           & .508$\pm$.012 & .590$\pm$.009 & .607$\pm$.010$^\dagger$  & +.099 \\
Gallbladder        & .518$\pm$.011 & .598$\pm$.009 & .614$\pm$.010$^\dagger$  & +.096 \\
Aorta              & .637$\pm$.008 & .723$\pm$.006 & .714$\pm$.007          & +.077 \\
Esophagus          & .497$\pm$.013 & .578$\pm$.010 & .593$\pm$.011$^\ddagger$ & +.096 \\
Stomach            & .614$\pm$.010 & .701$\pm$.008 & .690$\pm$.008          & +.076 \\
Adrenal (L/R)      & .483$\pm$.015 & .567$\pm$.012 & .586$\pm$.012$^\ddagger$ & +.103 \\
Lung lobes (agg.)  & .682$\pm$.006 & .768$\pm$.005 & .757$\pm$.005          & +.075 \\
Heart LV myo       & .583$\pm$.011 & .672$\pm$.009 & .662$\pm$.009          & +.079 \\
Heart LV blood     & .590$\pm$.010 & .681$\pm$.008 & .673$\pm$.008          & +.083 \\
IVC                & .608$\pm$.011 & .696$\pm$.009 & .686$\pm$.009          & +.078 \\
Colon              & .595$\pm$.012 & .682$\pm$.010 & .671$\pm$.010          & +.076 \\
Duodenum           & .531$\pm$.013 & .618$\pm$.010 & .633$\pm$.011$^\ddagger$ & +.102 \\
Femur (L/R)        & .682$\pm$.005 & .772$\pm$.004 & .761$\pm$.005          & +.079 \\
Brain              & .655$\pm$.006 & .744$\pm$.005 & .733$\pm$.005          & +.078 \\
Trachea            & .669$\pm$.006 & .756$\pm$.005 & .745$\pm$.005          & +.076 \\
\midrule
Mean of 18 rows    & .608          & .694          & .691                    & +.084 \\
\bottomrule
\end{tabular}
\end{table}

\begin{figure}[!t]
\centering
\includegraphics[width=\columnwidth]{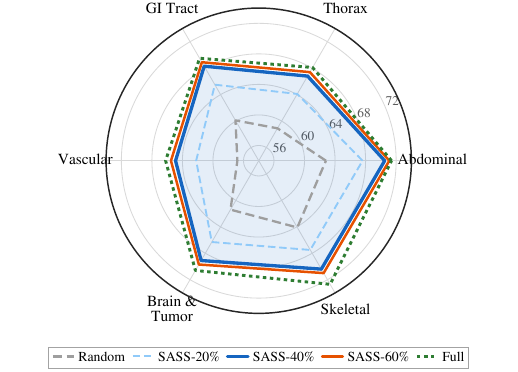}
\caption{Category-level budget comparison. SASS-40\% closely tracks Full-100\% across all six categories while improving substantially over Random and SASS-20\%.}
\label{fig:budget_radar}
\end{figure}

Table~\ref{tab:high_priority} summarizes 18 high-priority reporting targets spanning major anatomical systems. Based on the unweighted mean of the displayed row means, SASS retains approximately 99.6\% of Full Dataset performance using 40\% of the annotations (0.691 vs.\ 0.694). Its mean Dice exceeds Full for the pancreas, gallbladder, adrenal glands, esophagus, and duodenum, while well-represented organs such as the liver and kidneys retain ${\sim}$98.5\% of their Full Dice. These results indicate that targeted long-tail reallocation largely preserves performance on well-represented structures.

\subsubsection{Structure-Level Performance}
\label{sec:hard_organs_analysis}

\begin{table*}[!t]
\centering
\caption{Per-structure Dice and bootstrap uncertainty for seven representative Hard structures (mean $\pm$ std, three runs). $n_s$: canonical structure categories per row after the mapping in Supplementary Table~S1; $\phi$: validation-feedback gap. $\dagger$ marks differences whose 95\% CI excludes zero; intervals are not adjusted for multiplicity across structures. $\ddagger$: positive mean difference whose interval includes zero.}
\label{tab:hard_organs}
\scriptsize
\setlength{\tabcolsep}{5pt}
\begin{tabular}{@{}lccccccc@{}}
\toprule
Structure & $n_s$ & Rand & Full & SASS & $\Delta$(S--F) & 95\% CI & $\phi$ \\
\midrule
Pancreas       & 1 & 0.508$\pm$0.012 & 0.590$\pm$0.009 & 0.607$\pm$0.010$^\dagger$  & +0.017 & $[0.004,0.030]$  & 0.72 \\
Adrenal (L/R)  & 1 & 0.483$\pm$0.015 & 0.567$\pm$0.012 & 0.586$\pm$0.012$^\ddagger$ & +0.019 & $[-0.001,0.039]$ & 0.81 \\
Esophagus      & 1 & 0.497$\pm$0.013 & 0.578$\pm$0.010 & 0.593$\pm$0.011$^\ddagger$ & +0.015 & $[-0.003,0.033]$ & 0.68 \\
Heart RV       & 1 & 0.418$\pm$0.017 & 0.496$\pm$0.014 & 0.507$\pm$0.014$^\ddagger$ & +0.011 & $[-0.008,0.030]$ & 0.89 \\
Heart LA       & 1 & 0.441$\pm$0.015 & 0.534$\pm$0.012 & 0.549$\pm$0.013$^\ddagger$ & +0.015 & $[-0.005,0.035]$ & 0.83 \\
Lung vessel    & 1 & 0.453$\pm$0.014 & 0.545$\pm$0.011 & 0.559$\pm$0.012$^\ddagger$ & +0.014 & $[-0.004,0.032]$ & 0.76 \\
Gallbladder    & 1 & 0.518$\pm$0.011 & 0.598$\pm$0.009 & 0.614$\pm$0.010$^\dagger$  & +0.016 & $[0.003,0.029]$  & 0.61 \\
\bottomrule
\end{tabular}
\end{table*}

Table~\ref{tab:hard_organs} reports per-structure Dice with bootstrap uncertainty. SASS yields positive mean differences over Full for all seven structures (1.1--1.9\,pp). The pancreas and gallbladder have confidence intervals excluding zero; adrenal glands and esophagus show positive mean differences whose intervals include zero. Heart chambers exhibit the largest feedback gaps ($\phi{=}0.83$--$0.89$) yet only moderate improvements, with high run-to-run variability ($\sigma{=}0.013$--$0.014$) consistent with cardiac motion and boundary ambiguity after $128^3$ resampling. This indicates that feedback amplification and accuracy gains need not scale proportionally across structures. Results aggregated over all 108 categories are reported in Table~\ref{tab:main_difficulty} and Table~\ref{tab:per_category}.

\subsubsection{Category-Level Performance}
\label{sec:per_category}

\begin{table*}[!t]
\centering
\caption{Per-category test Dice at epoch~300 (mean $\pm$ std, three runs). Subset methods use 40\%; Full and Full+CB use 100\%. Bold: best 40\%-budget result.}
\label{tab:per_category}
\scriptsize
\setlength{\tabcolsep}{3pt}
\begin{tabular}{@{}lcccccccccc@{}}
\toprule
Category & Random & Full & Full+CB & Entropy & Coreset & BADGE & LESS & LESS+O & SASS & $\Delta$(S--R) \\
\midrule
A: Abdominal & 0.628$\pm$0.004 & 0.714$\pm$0.003 & 0.716$\pm$0.004 & 0.643$\pm$0.004 & 0.639$\pm$0.004 & 0.651$\pm$0.004 & 0.654$\pm$0.004 & 0.675$\pm$0.003 & \textbf{0.705$\pm$0.004} & +0.077 \\
B: Thorax    & 0.589$\pm$0.004 & 0.681$\pm$0.003 & 0.684$\pm$0.004 & 0.603$\pm$0.004 & 0.599$\pm$0.004 & 0.611$\pm$0.004 & 0.612$\pm$0.004 & 0.635$\pm$0.004 & \textbf{0.668$\pm$0.004} & +0.079 \\
C: GI Tract  & 0.601$\pm$0.005 & 0.695$\pm$0.004 & 0.698$\pm$0.004 & 0.616$\pm$0.005 & 0.612$\pm$0.005 & 0.624$\pm$0.004 & 0.626$\pm$0.004 & 0.649$\pm$0.004 & \textbf{0.683$\pm$0.004} & +0.082 \\
D: Vascular  & 0.568$\pm$0.005 & 0.662$\pm$0.004 & 0.666$\pm$0.005 & 0.583$\pm$0.005 & 0.578$\pm$0.005 & 0.591$\pm$0.005 & 0.594$\pm$0.005 & 0.619$\pm$0.005 & \textbf{0.649$\pm$0.005} & +0.081 \\
E: Brain\,\&\,Tumor & 0.614$\pm$0.005 & 0.706$\pm$0.004 & 0.709$\pm$0.005 & 0.628$\pm$0.005 & 0.624$\pm$0.005 & 0.636$\pm$0.005 & 0.639$\pm$0.005 & 0.657$\pm$0.005 & \textbf{0.691$\pm$0.004} & +0.077 \\
F: Skeletal  & 0.641$\pm$0.003 & 0.727$\pm$0.003 & 0.726$\pm$0.003 & 0.652$\pm$0.004 & 0.649$\pm$0.004 & 0.659$\pm$0.003 & 0.663$\pm$0.003 & 0.677$\pm$0.003 & \textbf{0.704$\pm$0.003} & +0.063 \\
\midrule
Overall      & 0.614$\pm$0.003 & 0.700$\pm$0.002 & 0.702$\pm$0.003 & 0.628$\pm$0.003 & 0.624$\pm$0.003 & 0.637$\pm$0.003 & 0.638$\pm$0.003 & 0.658$\pm$0.003 & \textbf{0.688$\pm$0.003} & +0.074 \\
\bottomrule
\end{tabular}
\end{table*}

Table~\ref{tab:per_category} and Fig.~\ref{fig:budget_radar} show that SASS leads among 40\%-budget methods in all six anatomical categories. Its improvement over Random ranges from 6.3\,pp (Skeletal) to 8.2\,pp (GI Tract), while retaining 96.8\% of Full on Skeletal despite the group-budget constraint. The progression from Random to LESS+Organ and then SASS reflects complementary gains from informative selection, category-level priors, and stage adaptation.

\subsubsection{Annotation-Budget Scaling and Training Dynamics}
\label{sec:budget_curve}

\begin{figure*}[!t]
\centering
\includegraphics[width=\textwidth]{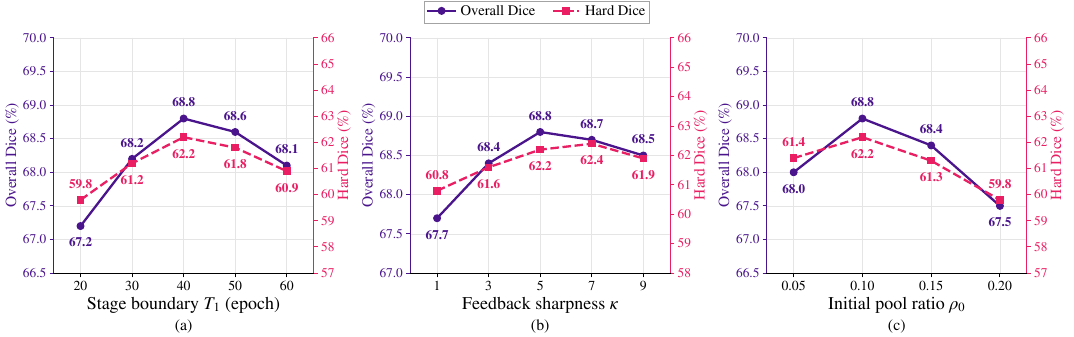}
\vspace{-4mm}
\caption{Hyperparameter sensitivity. Overall and Hard-structure Dice remain stable around the adopted configuration (dashed lines) for (a) stage boundary $T_1$, (b) feedback sharpness $\kappa$, and (c) initial pool ratio $\rho_0$.}
\label{fig:hyperparam_sensitivity}
\end{figure*}

\begin{table}[!t]
\centering
\caption{Annotation-budget scaling on the held-out test set (epoch~300, 10-click). Recovery is relative to Full overall Dice.}
\label{tab:budget_scaling_test}
\small
\setlength{\tabcolsep}{3.5pt}
\begin{tabular}{@{}ccccccc@{}}
\toprule
$\rho$ & Samples & Overall & Easy & Med. & Hard & Recovery \\
\midrule
0.20 & ${\sim}$14k & 0.660 & 0.692 & 0.645 & 0.582 & 94.3\% \\
\textbf{0.40} & \textbf{28k} & \textbf{0.688} & \textbf{0.715} & \textbf{0.675} & \textbf{0.622} & \textbf{98.3\%} \\
0.60 & ${\sim}$42k & 0.694 & 0.721 & 0.682 & 0.626 & 99.1\% \\
1.00 & 70{,}351 & 0.700 & 0.733 & 0.690 & 0.610 & 100\% \\
\bottomrule
\end{tabular}
\end{table}

Table~\ref{tab:budget_scaling_test} supports $\rho{=}0.40$ as the principal operating point. Three observations emerge. First, SASS-20\% already recovers 94.3\% of Full performance, indicating that stage-adaptive selection remains effective even when annotation is one-fifth of the full pool. Second, increasing the budget from 40\% to 60\% yields only 0.6\,pp, consistent with a data-saturation regime. Third, SASS-20\% still substantially exceeds Random-40\% despite using half the annotation budget, providing a striking demonstration that selection quality can compensate for selection quantity.

The complete training-dynamics and curriculum-stability analysis is provided in Supplementary Fig.~S1. SASS separates from the active-learning baselines during the gradient-enabled stages, while the forgetting-guided curriculum primarily improves post-expansion recovery rather than endpoint Dice.

\begin{table}[!t]
\centering
\caption{Volume-disjoint robustness check. All 40\%-budget methods use the same annotation budget.}
\label{tab:volume_disjoint}
\footnotesize
\setlength{\tabcolsep}{5pt}
\begin{tabular}{@{}lcccc@{}}
\toprule
Method & Budget & Overall & Medium & Hard \\
\midrule
Full Dataset & 100\% & 0.687 & 0.677 & 0.596 \\
Random       & 40\%  & 0.602 & 0.583 & 0.512 \\
LESS+Organ   & 40\%  & 0.647 & 0.629 & 0.563 \\
SASS         & 40\%  & 0.675 & 0.663 & 0.604 \\
\bottomrule
\end{tabular}
\end{table}

Under the stricter volume-disjoint split (Table~\ref{tab:volume_disjoint}), SASS has the highest reported overall mean among the evaluated 40\%-budget methods and a higher Hard-group mean than Full (0.604 vs.\ 0.596). These point estimates are consistent with the principal performance pattern under the grouped split.

\subsection{Ablation and Mechanism Analysis}
\label{sec:ablation}

\subsubsection{Component Contributions}
\label{sec:component_ablation}

\begin{figure}[!t]
\centering
\includegraphics[width=0.97\columnwidth]{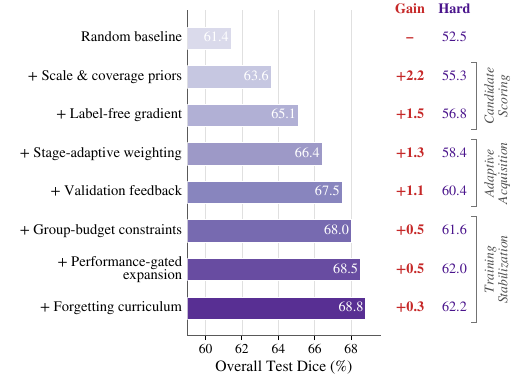}
\caption{Cumulative ablation at 40\% budget. Bars: overall test Dice; red values: incremental gains; purple: Hard-structure Dice. Components grouped into scoring, adaptive acquisition, and stabilization.}
\label{fig:cumulative_ablation}
\end{figure}

Figure~\ref{fig:cumulative_ablation} reports a cumulative ablation in which components are added to the Random baseline in a fixed order. The overall Dice values and incremental gains are provided in Supplementary Table~S3; the increments quantify gains within this cumulative configuration rather than order-independent effects.

Adding the scale-aware and coverage-balance priors raises Dice from 0.614 to 0.636 (+2.2\,pp), supporting category-level structure as a useful acquisition cue. Incorporating the label-free gradient score then increases Dice to 0.651 (+1.5\,pp), indicating that model-dependent knowledge-evolution signals provide complementary information beyond the two priors. The next two components make acquisition responsive to training dynamics: stage-adaptive weighting contributes 1.3\,pp, supporting the use of stage-dependent scoring as training progresses, and validation-driven feedback adds a further 1.1\,pp, indicating that category-level performance gaps provide useful signals for updating acquisition priorities. The remaining components target selection balance and training stability: group-budget constraints add 0.5\,pp by limiting budget concentration in repetitive categories, the performance-gated expansion schedule contributes 0.5\,pp while allowing the model to adapt between query batches (all configurations in this ablation already use multi-round acquisition, so this increment isolates expansion-interval adaptivity rather than multi-round acquisition itself), and the forgetting-guided curriculum adds 0.3\,pp.

\begin{figure}[!t]
\centering
\includegraphics[width=\columnwidth]{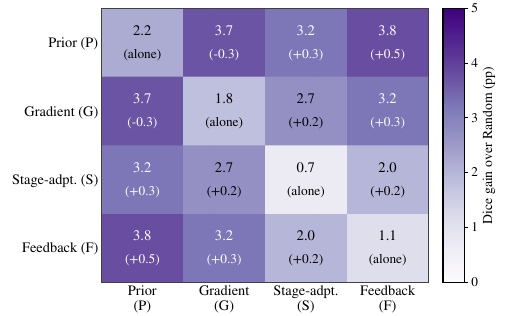}
\caption{Component interaction analysis. Each cell: Dice gain over Random for the corresponding pair; diagonal: individual components. Parentheses report the interaction beyond additivity, computed as the paired gain minus the sum of the two individual gains.}
\label{fig:component_heatmap}
\end{figure}

Figure~\ref{fig:component_heatmap} complements the cumulative ablation by showing pairwise interactions. Prior and gradient scoring provide the largest individual gains, while stage adaptation and validation feedback produce additional synergy when combined with the scoring components. This confirms that the main components are complementary rather than interchangeable.

\subsubsection{Sensitivity Analysis}
\label{sec:sensitivity}

The sensitivity analyses are provided in Supplementary Tables~S5--S7. The adopted Stage~3 weighting ($\alpha_1{=}0.70$) preserves a nonzero prior contribution, the group cap ($\eta{=}0.15$) prevents skeletal over-concentration, and validation-triggered rescoring gives the best Dice with moderate overhead (4.7\%).

For context, the 4.2\,pp gain from one-shot to validation-triggered rescoring is comparable in magnitude to the 4.4\,pp spread from Random to the strongest non-SASS 40\%-budget selector in Table~\ref{tab:main_difficulty} (0.614--0.658). This descriptive comparison indicates that acquisition timing can affect performance on the same scale as the choice of selection criterion.

Figure~\ref{fig:hyperparam_sensitivity} shows smooth performance variation across the tested values of $T_1$, $\kappa$, and $\rho_0$, with the adopted configuration achieving the highest overall Dice in each sweep. Here, $T_1$ denotes the stage-transition boundary defined in Eq.~\ref{eq:activation}, $T_2$ specifies the end of the interpolation interval in Eq.~\ref{eq:stage2_interp}, $\kappa$ is the feedback-sharpness parameter in Eq.~\ref{eq:gap}, and $\rho_0$ is the initial-pool ratio defined in Eq.~\ref{eq:initial_pool}. For the stage boundary $T_1$ (Fig.~\ref{fig:hyperparam_sensitivity}(a)), performance improves as the transition from prior-guided to gradient-enabled acquisition is delayed from 20 to 40 epochs, reaching 68.8\% Overall Dice and 62.2\% Hard Dice at $T_1=40$, before gradually declining for later transitions. This suggests that activating gradient-based evidence too early, before sufficiently mature representations have formed, is less effective, whereas excessive delay also limits its contribution. For feedback sharpness $\kappa$ (Fig.~\ref{fig:hyperparam_sensitivity}(b)), performance remains relatively stable over a broad range, with Overall Dice peaking at 68.8\% for $\kappa=5$ and Hard Dice varying only moderately, indicating limited sensitivity to the precise feedback strength. The initial pool ratio $\rho_0$ (Fig.~\ref{fig:hyperparam_sensitivity}(c)) exhibits a similarly smooth trend, with $\rho_0=0.10$ providing the best balance between Overall Dice (68.8\%) and Hard Dice (62.2\%); both smaller and larger initial pools reduce overall performance.

\subsubsection{Dynamic Acquisition}
\label{sec:dynamic_acquisition}

The validation-triggered expansion process improves validation performance as the labeled pool grows, with the 40\% target reached shortly after the Stage~3 transition. The representative trajectory is provided in Supplementary Table~S4; realized expansion epochs may vary across runs.

\subsubsection{Selection Evolution}
\label{sec:selection_evolution}

\begin{figure*}[!t]
\centering
\includegraphics[width=\textwidth]{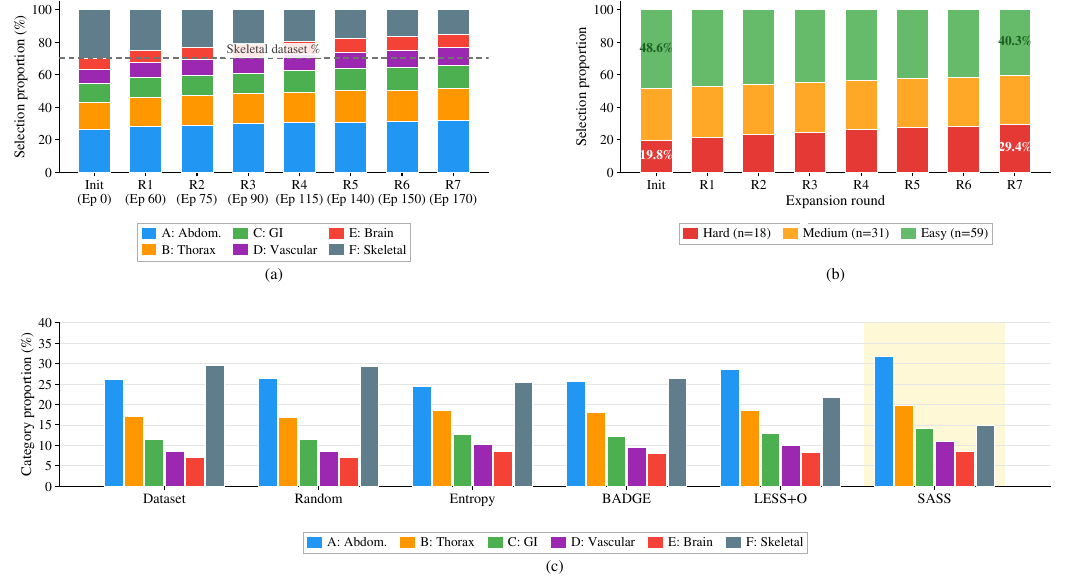}
\caption{Selection evolution. (a) Category distribution per query batch; dashed line: Skeletal candidate-pool share (29.5\%). (b) Difficulty-group distribution: Hard share rises from 19.8\% to 29.4\% ($2.1{\times}$ candidate-pool proportion). (c) Final cumulative pool by method: SASS reduces Skeletal to 14.8\%, shifting budget toward Abdominal (${\sim}$32\%) and GI (${\sim}$14\%).}
\label{fig:selection_evolution}
\end{figure*}

Having quantified what SASS achieves, we now examine \emph{how} annotation budget is allocated over time. Figure~\ref{fig:selection_evolution} visualizes selection behavior from three perspectives.

In Stage~1 (Fig.~\ref{fig:selection_evolution}(a)), category-level priors and group constraints distribute the budget broadly while applying the per-group caps to ribs and vertebrae. During Stage~2, as the gradient score receives increasing weight, the Abdominal share rises from 28.5\% to 31.2\%. In Stage~3 ($\alpha_1{=}0.70$), the Skeletal category allocation drops to 8.5\% in the final query batch and the released budget is redistributed across the remaining categories.

Figure~\ref{fig:selection_evolution}(b) shows a corresponding difficulty-group shift: the Hard-structure share increases from 19.8\% to 29.4\%---approximately $2.1{\times}$ its candidate-pool proportion of 13.8\%---providing higher annotation density for difficult structures and explaining the group-level less-is-more pattern in Table~\ref{tab:main_difficulty}.

The final-pool comparison (Fig.~\ref{fig:selection_evolution}(c)) shows that Random, Entropy, and BADGE remain largely dataset-proportional, whereas SASS produces the strongest departure from the candidate-pool distribution while preserving broad anatomical coverage. These shifts are consistent with the combined effects of category-level priors, group constraints, stage-adaptive weighting, and validation feedback, providing an operational counterpart to the performance gains reported above.

\subsection{Discussion and Analysis}
\label{sec:discussion_analysis}
\label{sec:discussion}
\label{sec:discussion_summary}
\label{sec:why_sass_works}
\label{sec:generalization}

Active learning is commonly framed as choosing which candidates to label under a fixed annotation budget. Our results show that, in long-tailed dense prediction, annotation efficiency also depends on how the budget is distributed across categories and when model-derived evidence is sufficiently reliable to guide acquisition. SASS addresses these requirements through prior-guided category allocation, label-free self-supervised gradient scoring, and validation-gated activation of that score. On a testbed of more than 100{,}000 samples spanning 108 anatomical structures, SASS recovers 98.3\% of full-dataset Dice at a 40\% training-pool annotation budget and outperforms BADGE by 5.1 percentage points. The group-level ``less-is-more'' result, together with structure-level differences whose confidence intervals exclude zero, further shows that targeted reallocation can improve underrepresented categories while retaining most overall performance.

SASS addresses a different form of non-stationarity from prior adaptive acquisition. RALF adjusts the utility of exploration and exploitation criteria that remain available throughout acquisition~\cite{ebert2012ralf}, whereas SASS determines whether model-derived gradient evidence is sufficiently reliable to enter selection at all. Imbalance-aware acquisition has established the value of balancing classes or regions~\cite{aggarwal2020active,cai2021revisiting}, but SASS makes long-tail allocation performance-responsive through validation feedback and maintains this control after gradient scoring begins. Unlike supervised gradient and influence estimators, which require candidate labels or construct surrogates for a supervised candidate loss~\cite{ash2020badge,wang2022boosting,xia2024less}, SASS derives its sample-level gradient score from a self-supervised teacher--student objective without candidate masks. Together, these differences define a reliability-gated decomposition of acquisition: category-level budget allocation remains active throughout training, whereas model-derived sample ranking is introduced only after the task-specific validation criterion is met.

In addition, our design has a training-dynamics rationale. In finite-width neural networks, the model's gradient geometry---that is, the relationships among parameter gradients---changes rapidly early in training and evolves more slowly thereafter~\cite{fort2020deep}; consequently, a ranking produced by an immature model may become stale by a later acquisition round. Excluding candidate gradients before the validation gate prevents such early rankings from shaping the pool; after the gate, gradient scoring supports sample-level refinement while the prior terms maintain category coverage. The validation criterion in Eq.~\ref{eq:activation} operationalizes signal readiness for the present task; it is not claimed to measure gradient-geometry stability directly or to define a universal phase boundary.

The empirical analyses support this reliability-gated interpretation. The cumulative ablation and pairwise interaction results show that category-level priors, label-free gradient scoring, stage adaptation, and validation feedback are complementary rather than interchangeable (Supplementary Table~S3 and Fig.~\ref{fig:component_heatmap}). More directly, validation-triggered rescoring improves one-shot selection by 4.2 percentage points (Supplementary Table~S7), showing that refreshing model-derived scores as training evolves materially affects performance. The selection-evolution analysis further shows that SASS shifts the labeled pool away from the candidate-pool distribution toward Hard and underrepresented categories (Fig.~\ref{fig:selection_evolution}), providing behavioral evidence that the performance gains are accompanied by the intended long-tail reallocation.

The larger gains for several small or difficult structures cannot be attributed to organ size alone. A more plausible explanation is that these structures remain poorly learned under Full training and therefore benefit more from targeted annotation than structures that are already near saturation. Consistent with this interpretation, the liver and kidneys retain approximately 98.5\% of their Full Dice at the 40\% SASS budget (Table~\ref{tab:high_priority}), while the Hard group exceeds the Full reference at both $\rho{=}0.40$ and $\rho{=}0.60$ (0.622 and 0.626 versus 0.610; Table~\ref{tab:budget_scaling_test}). These results attribute the ``less-is-more'' effect to more effective annotation allocation: SASS preserves sufficient supervision for well-learned categories while reallocating annotation effort toward difficult and under-represented categories, where additional samples provide greater benefit. This enables improved performance on the Hard group despite using substantially fewer annotations overall.

The contrast with Full+CB in Section~\ref{sec:al_comparison} indicates that training-time class rebalancing and acquisition-time pool construction address different objectives: the former improves learning from an already labeled dataset, whereas the latter reallocates a limited annotation budget toward underrepresented categories. SASS therefore does not uniformly replace class-balanced full-data training; its advantage lies in retaining most overall performance under a restricted annotation budget while improving the long tail.

Although instantiated on the point-prompt 3D segmentation architecture of Wang et al.~\cite{wang2023sam_med3d}, SASS operates at the data-selection level and does not require structural changes to the segmentation network. Its acquisition components use standard model forward/backward operations, candidate category identifiers available before mask annotation, and feedback derived from the validation set, rather than architecture-specific modules. SASS could therefore in principle be adapted to other dense-prediction systems, including self-configuring CNN pipelines such as nnU-Net~\cite{isensee2021nnu} and hybrid transformer architectures such as TransUNet~\cite{chen2024transunet}; empirical validation across architectures remains future work.

Some structures remain difficult under all evaluated selection strategies. The relatively low Dice of heart cavities and fine pulmonary vasculature is consistent with motion, boundary ambiguity, and limited spatial detail after $128^3$ resampling. SASS nevertheless improves their mean Dice over Random selection, while validation feedback assigns greater acquisition priority to these underperforming categories. Thus, targeted reallocation can reduce under-representation, although it does not eliminate limitations associated with the imaging and preprocessing setting.

\subsection{Limitations}
\label{sec:limitations}

Several limitations also motivate further development of SASS. First, the absolute Dice values depend on the point-prompt and resampling protocol adopted in this study and may vary with alternative prompting strategies or higher-resolution training. Nevertheless, all evaluated selection methods share the same architecture, prompting protocol, and resolution, enabling controlled comparison of their relative data-selection effectiveness. Second, acquisition is simulated from a fully annotated pool, following standard active-learning practice~\cite{settles2009active}. Extending SASS to prospective expert-in-the-loop acquisition will enable evaluation of annotation time, interaction cost, and workflow efficiency in realistic clinical settings. Third, the prior-guided components in our medical instantiation exploit category identifiers available before dense mask annotation through the dataset organization. In applications where such metadata are unavailable, these components can be instantiated with alternative task-appropriate priors, while the label-free gradient scorer remains directly applicable. Finally, gradient-enabled acquisition introduces additional computation for candidate-level gradient extraction. The stage-adaptive design already mitigates this cost by bypassing gradient computation during Stage 1 and activating it only when model-derived signals become sufficiently informative; further improvements in gradient approximation and candidate screening could enhance scalability to even larger pools.

Overall, these considerations identify natural directions for extending SASS beyond the present setting. Within the controlled evaluation considered here, all methods are compared under matched architectures, prompting protocols, and resolutions, with subset-selection methods operating under the same annotation budget. The results therefore provide a consistent assessment of the benefits of stage-adaptive, label-free, and category-aware sample selection.

\section{Conclusion and Future Work}
\label{sec:conclusion}

We presented SASS, a stage-adaptive data selection framework for annotation-efficient dense prediction under long-tailed category distributions. Under a substantially reduced annotation budget, SASS approaches full-dataset performance while surpassing it on the hardest categories, and outperforms established active-learning acquisition baselines evaluated here. These findings show that targeted labeled-pool construction can preserve most overall performance while improving performance in the long tail.

Beyond ranking individual samples, SASS also determines how the annotation budget is distributed across long-tailed categories and when model-derived scores should begin to guide acquisition. Because this formulation operates at the data-selection level without modifying the underlying architecture, it has the potential to extend to other dense-prediction systems and long-tailed domains. Future work should quantify expert annotation savings in prospective workflows and reduce candidate-level scoring cost for larger unlabeled pools. More broadly, deciding when model-derived scores are reliable enough to guide selection may inform data curation for large-scale pretraining and instruction tuning.

\section*{Data Availability}
This study uses only publicly available, de-identified 3D medical image datasets and generates no new patient data. Code, training pipelines, and evaluation scripts are available for peer review at \url{https://anonymous.4open.science/r/SASS-CD18} and will be publicly released under an open-source license upon acceptance.

\section*{Conflict of Interest}
The authors declare no competing interests.

\bibliographystyle{IEEEtran}
\bibliography{references}

\end{document}